\documentclass[10pt,twocolumn,letterpaper]{article}

\usepackage[pagenumbers]{cvpr} % arXiv: non-anonymous with page numbers
\usepackage{algorithm}
\usepackage{algpseudocode}

\usepackage{graphicx}
\usepackage{amsmath,amssymb}
\usepackage{booktabs,tabularx,array,adjustbox}
\usepackage{xcolor}
\usepackage{etoolbox}
\usepackage{caption}
\usepackage{subcaption}
\usepackage{float}
\usepackage{placeins}
\usepackage{microtype}

\patchcmd{\abstract}{\vspace*{12pt}}{\vspace*{6pt}}{}{}

\newcommand{\suppOcclusionPlot}[2]{%
    \IfFileExists{sec/figures/suppl_fig/#2}{%
        \includegraphics[width=#1,keepaspectratio]{sec/figures/suppl_fig/#2}%
    }{%
        \IfFileExists{figures/#2}{%
            \includegraphics[width=#1,keepaspectratio]{figures/#2}%
        }{%
            \fbox{\begin{minipage}[c][1.0in][c]{0.92\linewidth}
                \centering\footnotesize
                Missing plot: \texttt{\detokenize{#2}}\\[4pt]
                Upload to \texttt{sec/figures/suppl\_fig/}
            \end{minipage}}%
        }%
    }%
}

\newcommand{\promptfont}{\fontencoding{T1}\sffamily\selectfont}
\newcommand{\promptstep}[2]{%
    \par\addvspace{6pt}{\fontsize{10.5}{12}\selectfont\bfseries #1\par}%
    \nobreak\noindent\textbf{Input}\enspace #2\par\smallskip}
\newcommand{\promptreturn}[2]{%
    \par\smallskip\noindent\textbf{#1}\enspace #2\par}

\definecolor{cvprblue}{rgb}{0.21,0.49,0.74}
\usepackage[pagebackref,breaklinks,colorlinks,allcolors=cvprblue]{hyperref}
\usepackage{subfiles}

\def\confName{CVPR}
\def\confYear{2027}

\newcommand{\SDTrackTitle}{SD-Track: Sparse-to-Dense Amodal Fusion for Question-Guided Tracking}
\title{\SDTrackTitle}

\author{
Minseon Park$^{1\dagger}$ \quad
Changwoo Jeon$^{2\dagger}$ \quad
Jian Kim$^{3\dagger}$ \quad
Inhong Paek$^{4\dagger}$\\
Jaeyun Shin$^{5}$ \quad
Jong Chul Ye$^{5*}$\\[4pt]
$^{1}$Sungkyunkwan University \quad
$^{2}$Yonsei University\\
$^{3}$Soongsil University \quad
$^{4}$UC Berkeley \quad
$^{5}$KAIST
}

\begin{document}
% Keep the main-paper typography unchanged.
\microtypesetup{protrusion=false,expansion=false}
\raggedbottom

% Preserve the existing title/teaser/abstract setup.
\input{sec/0_abstract}
%\maketitle
% \begingroup
% \renewcommand{\thefootnote}{\fnsymbol{footnote}}
% \begin{NoHyper}
% \footnotetext[2]{Equal contribution.}
% \footnotetext[1]{Corresponding author:
% \texttt{jong.ye@kaist.ac.kr}.}
% \end{NoHyper}
% \endgroup

\section{Introduction}
\label{sec:introduction}

Spatio-temporal video grounding (STVG) aims to localize objects in space and time from natural-language queries. These queries can be referring expressions~\cite{Wu_2023_CVPR,lu2026storm,Zhang2020VidSTG,Tang2021HCSTVG} or questions that require identifying relevant instances from video events~\cite{Zhang2020VidSTG,Patraucean2023PerceptionTest,Hannan2025SVAG}. Spatio-temporal localization and tracking support applications such as autonomous driving~\cite{Yu2020BDD100K,Zhang2019mmMOT,Weng2020AB3DMOT}, traffic monitoring~\cite{Tang2019CityFlow,Shim2021CityTraffic,Li2022AICity}, sports analytics~\cite{cui2023sportsmot,Cioppa2022SoccerNet,Scott2024TeamTrack}, robot perception~\cite{Martin2019JRDB,VendrowEtAl2023JRDBPose}, etc.

\begin{figure*}[!t]
    \centering
    \includegraphics[width=\textwidth]{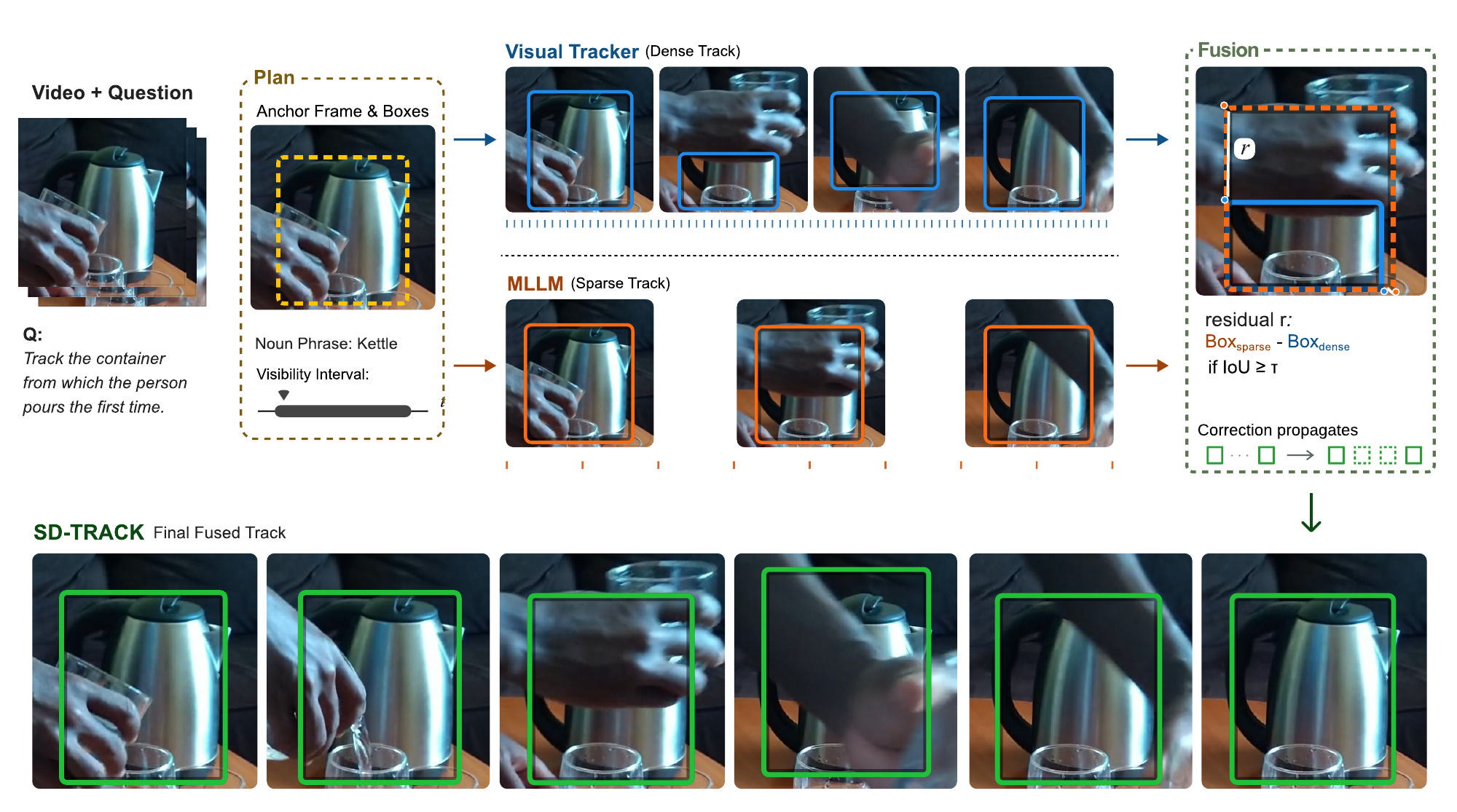}
    \caption{\textbf{Overview of SD-Track.}
    Given a video and a question, an MLLM (Gemini 3.7 Flash) identifies the target objects and constructs a shared tracking plan containing instance identities, anchor boxes (yellow), and visibility intervals.     Guided by this plan, a visual tracker (SAM3) produces dense trajectories of the visible extent (blue, $b^D$), while the MLLM estimates sparse amodal bounding boxes (orange, $b^A$). Reliability-aware fusion selects reference frames where $\operatorname{IoU}(b^A, b^D) \geq \tau$ and propagates coordinate residuals $r = b^A - b^D$ along dense trajectories. The bottom row shows the resulting fused track (green) across video frames.}
    \label{fig:qtrace_overview}
\end{figure*}

Existing STVG approaches either learn grounding and tracking jointly~\cite{Wu_2023_CVPR,lu2026storm} or employ MLLMs to guide pretrained visual models~\cite{chamiti2025refergpt,huang2024unleashing,hu2026yestrack,zhao2026agentic}. However, tracking query-relevant objects becomes challenging when they are partially or fully occluded by other objects. In such cases, the tracker must preserve the target's identity and estimate its full spatial extent despite limited visibility. Conventional localization based on visible regions may produce bounding boxes that shrink under occlusion, leading to inconsistent object trajectories.
Amodal localization estimates the full extent of an object, including occluded regions, while temporal reasoning is needed to distinguish temporary occlusion from departure.
 Although MLLMs can reason about occluded objects using video context, applying them for frame-by-frame amodal localization incurs substantial computational or API costs. This motivates efficient approaches that maintain accurate amodal localization and consistent tracking throughout the video.

To address these challenges, we propose \textbf{SD-Track}, a question-guided tracking framework that combines sparse amodal localization with dense visual tracking.
Given a question, an MLLM identifies the target objects and generates a shared tracking plan to guide both components.
Following this plan, SAM3~\cite{qtrace_sam3} produces dense visual tracks, while the MLLM estimates amodal bounding boxes at sparsely sampled frames, capturing the full extent of each target even under occlusion.
By relying on visual tracking between these sparse MLLM observations, SD-Track substantially reduces the API cost compared with frame-by-frame MLLM-based amodal localization.

In particular, to combine these complementary predictions, we introduce a sparse-to-dense amodal fusion mechanism
(\cref{fig:qtrace_overview}).
Dense visual tracks provide continuous estimates of object motion, while sparse amodal boxes correct the visible-region estimates to account for occluded regions.
These corrections are propagated along the visual trajectories, enabling amodal localization at intermediate frames without additional MLLM calls.
When visual predictions are unavailable, the sparse MLLM predictions are retained to preserve the available amodal observations.
Our contributions are:
\begin{itemize}
    \item We introduce SD-Track, a framework for amodal tracking from natural-language queries that coordinates sparse MLLM localization and dense visual tracking through a shared tracking plan.
    \item We propose a sparse-dense fusion strategy that corrects visual trajectories using spatially consistent amodal observations and retains sparse predictions where visual tracking is unavailable.
    \item We evaluate SD-Track on Perception Test Grounded VQA, SVAG-Bench, VidSTG, and HC-STVG without benchmark-specific training. An accuracy and MLLM API cost analysis on the Perception Test validation subset measures the trade-off between dense MLLM-based amodal localization and sparse-dense fusion.
\end{itemize}

\section{Related Work}
\label{sec:related}

\paragraph{Spatio-Temporal Video Grounding.}
\label{sec:rw_grounding}
Spatio-Temporal Video Grounding (STVG) aims to localize objects referred to by natural language queries across video frames, as studied in VidSTG~\cite{Zhang2020VidSTG} and HC-STVG~\cite{Tang2021HCSTVG}. Referring multi-object tracking (RMOT)~\cite{Wu_2023_CVPR} extends this setting to multiple instances, while STORM~\cite{lu2026storm} integrates grounding and tracking in an end-to-end framework. Beyond appearance-based descriptions, queries may specify targets through actions, interactions, or temporal events, as reflected in VidSTG~\cite{Zhang2020VidSTG}, SVAG-Bench~\cite{Hannan2025SVAG}, and the Perception Test~\cite{Patraucean2023PerceptionTest}. SD-Track addresses these STVG tasks by combining sparse amodal localization with dense tracking in a zero-shot agentic framework.

\paragraph{MLLM-Guided Video Grounding.}
\label{sec:rw_mllm}
Recent approaches use multimodal large language models (MLLMs) to guide visual grounding. AL-Ref-SAM2~\cite{huang2024unleashing} selects initialization frames and boxes for SAM~2~\cite{ravi2024sam2}, while ReferGPT~\cite{chamiti2025refergpt} and YesTrack~\cite{hu2026yestrack} use MLLMs for target matching and track verification, respectively. ASTG~\cite{zhao2026agentic} employs collaborative reasoning agents for spatio-temporal localization. In many modular methods, MLLM reasoning primarily supports target selection or verification, while visual models provide dense localization. SD-Track extends this reasoning to spatial prediction, using sparse amodal estimates to refine dense trajectories.

\paragraph{Occlusion and Amodal Perception.}
\label{sec:rw_amodal}
Promptable video segmenters such as SAM~2~\cite{ravi2024sam2} and SAM~3~\cite{qtrace_sam3} propagate visible masks, which may shrink or disappear under occlusion. SAMURAI~\cite{yang2024samurai} and DAM4SAM~\cite{videnovic2025dam4sam} improve tracking robustness but do not explicitly estimate occluded object extent. Amodal perception instead estimates the full object extent, including hidden regions. This representation supports reasoning about object size and continued presence when visibility is limited. SAIL-VOS~\cite{hu2019sailvos} provides amodal video annotations to support occlusion reasoning and object-size prediction. TAO-Amodal~\cite{hsieh2023taoamodal} studies full-object tracking through partial and complete occlusion, motivated by applications in motion planning and navigation. Diffusion-VAS~\cite{chen2025diffusionvas} and Amodal SAM~\cite{zhang2026amodalsam} explore amodal prediction. SD-Track applies amodal extent estimation to query-guided localization by refining dense visual trajectories with sparse MLLM predictions.
\section{Method}
\label{sec:methodology}

We formulate the task and outline SD-Track in Sec.~\ref{sec:formulation}. Sections~\ref{sec:plan}, \ref{sec:localization}, \ref{sec:fusion} describe shared planning, sparse and dense localization, and trajectory fusion, respectively.

\subsection{Problem Formulation and Overview}
\label{sec:formulation}

Given a video $V=\{I_t\}_{t=0}^{T-1}$ of $T$ frames and a natural-language query $q$, our goal is to identify the referred object instances and predict their bounding-box trajectories.
The output is $\hat{\mathcal{Y}}=\{\hat{\tau}_k\}_{k=1}^{K}$, where $K$ is the number of identified target instances and
$\hat{\tau}_k=\{(t,\hat{b}_{k,t}) \mid t\in\hat{\mathcal{T}}_k\}$
denotes the predicted trajectory of instance $k$ over the frame indices $\hat{\mathcal{T}}_k\subseteq\{0,\ldots,T-1\}$.
Each bounding box $b=[x_1,y_1,x_2,y_2]^\top\in[0,1]^4$ is represented by normalized corner coordinates, and the identifier $k$ refers to the same physical instance throughout the video.
For occluded targets, we aim to predict amodal boxes that cover the full object extent, including hidden regions.

SD-Track first uses video-level reasoning to construct a shared tracking plan $P$ containing the target instance identifiers and their anchor boxes (Fig.~\ref{fig:qtrace_overview}).
This plan guides two complementary branches: sparse MLLM-based amodal localization and dense visual tracking.
We denote their predictions by $A=\{A_k\}_{k=1}^{K}$ and $D=\{D_k\}_{k=1}^{K}$, respectively, where $A_k$ and $D_k$ contain the predicted boxes and frame indices for instance $k$.
A fusion module $\Phi$ combines these predictions using the shared instance identifiers and propagates coordinate residuals from spatially consistent box pairs along the dense trajectories.
When dense predictions are unavailable, the corresponding sparse predictions are retained.

Let $G$ denote the MLLM and $S$ the visual tracker.
The overall pipeline is
\begin{align}
    P &= G_{\mathrm{plan}}(V,q), \nonumber\\
    A &= G_{\mathrm{sparse}}(V,P), \quad D=S(V,P), \nonumber\\
    \hat{\mathcal{Y}} &= \Phi(A,D,P).
    \label{eq:pipeline}
\end{align}
We use Gemini 3.7 Flash~\cite{qtrace_gemini_flash} for $G$ and SAM3~\cite{qtrace_sam3} for $S$.

\subsection{Shared Question-Guided Tracking Plan}
\label{sec:plan}

We first use video-level context to identify the physical instances referred to by the query. This may require reasoning about interactions, appearance order, or repeated participation in an event. We encode the selected instances in a shared tracking plan:
\begin{equation}
    P=\{(k,d_k,\mathcal{V}_k,\mathcal{U}_k)\}_{k=1}^{K},
    \label{eq:plan}
\end{equation}
where $k$ is an instance identifier, $d_k$ records distinguishing appearance cues, and $\mathcal{V}_k$ contains estimated visibility intervals. The set $\mathcal{U}_k=\{(u,p_{k,u})\}$ specifies selected source-frame indices $u$ and initialization boxes $p_{k,u}$ covering the visible extent of instance $k$. Visibility intervals indicate when a target is visually observed.

As illustrated in Fig.~\ref{fig:qtrace_overview}, the MLLM first produces a visual description of the queried object. A separate critic call checks the answer from the video and query, revises the description when needed, and provides an estimated target count as a hint for multi-instance planning. The anchor planner assigns persistent instance IDs, records appearance cues and visibility intervals, and prioritizes a common frame where all targets are visible, separated, and minimally occluded. If no suitable common frame is found, we request the smallest set up to three frames covering all target identities. Each selected source image is then provided to the MLLM with the video context and identity to predict initialization boxes for exactly the requested IDs.

We construct a uniformly sampled planning proxy with each frame labeled by its original source frame index. Off-grid anchor selections are snapped to the nearest permitted anchor frame. Selections aligned to the same frame are merged by uniting their target-ID lists, after which we recheck that every target is covered by at least one anchor. Each stage uses structured function outputs. For anchor boxes, we require an exact match to the requested ID set, coordinates within bounds, and positive area. The validated plan is saved and reused by both localization branches.

\subsection{Complementary Sparse and Dense Localization}
\label{sec:localization}

\paragraph{Dense visual tracking.}
We initialize SAM3 from the shared tracking plan. Each anchor box is encoded as two points at its top-left and bottom-right corners. SAM3 uses box-corner labels to interpret these points as a box prompt. All targets share one tracking session  and retain their planned instance identifiers.

We first select regularly spaced source frames. We then add all anchor frames and additional frames near the boundaries of the planned visibility intervals. After registering the anchor prompts, we propagate masks forward and backward over this frame sequence.

SAM3 predicts instance masks at the sampled frames. Empty masks are discarded. For each remaining mask, SAM3 returns its enclosing bounding box. We express these boxes in normalized corner coordinates and associate each prediction with its original source-frame index. The resulting observations form $D_k=\{(t,b^D_{k,t}):t\in\mathcal{T}^D_k\}$.

We then apply visibility gating to each instance separately. Predictions are retained within its planned visibility intervals, expanded by a temporal margin. This gate is intended to reduce drift onto distractors after the target disappears. Outside these intervals, the sparse branch can still estimate an occluded target's box when temporal context supports its continued presence.

\paragraph{Sparse amodal localization.}
The sparse branch uses the same tracking plan to localize targets on a coarser source-frame grid.
Each MLLM call receives up to eight sampled images and a compact version of the plan. The plan provides instance identifiers, appearance descriptions, visibility intervals, and anchor boxes.

The prompt requests a separate box for each target under its plan identifier. Each box represents one physical instance; occluders and neighboring objects are not treated as part of that instance. For partially occluded targets, the prompt asks for boxes covering the full extent, including hidden regions. For fully occluded targets, it asks the model to infer position and size from the temporal context in the sampled images. An instance should be omitted only when it has left the scene, rather than merely become occluded.

We validate the returned frame indices and instance identifiers. We also check identifier uniqueness, coordinate ranges, and positive box area. The validated predictions form $A_k=\{(t,b^A_{k,t}):t\in\mathcal{T}^A_k\}$. Sampling settings are provided in Sec.~\ref{sec:experimental_setup}.

\subsection{Reliability-Aware Amodal Fusion}
\label{sec:fusion}

Our key idea is to decouple temporal object motion from amodal extent estimation. Dense visual tracks provide  motion estimates, while sparse MLLM predictions provide corrections to the visible-region boxes. Rather than interpolating sparse amodal boxes directly, we propagate these corrections along the visual trajectories, preserving their temporal motion while recovering occluded object extent.

As in Fig.~\ref{fig:qtrace-sequences}, dense visual tracking follows the target across closely spaced frames, helping maintain its identity among similar-looking objects. However, boxes derived from visible masks can underestimate the target's extent under occlusion. Sparse amodal localization estimates the full extent, including hidden regions, but provides fewer observations and may confuse target identities. Our fusion module therefore uses the visual track as the base trajectory and corrects its coordinates with spatially consistent amodal boxes.

\paragraph{Residual estimation.}
We match predictions by their shared plan identifier (k) and select \textit{correction frames}, where both branches provide a box and their IoU is at least $\tau=0.30$. At each correction frame, we compute the coordinate residual from the visual box to the amodal box:
\begin{align}
\mathcal{R}_k
&=
\left\{
s \in \mathcal{T}^A_k \cap \mathcal{T}^D_k :
\operatorname{IoU}(b^A_{k,s}, b^D_{k,s}) \geq \tau
\right\},
\nonumber\\
r_{k,s}
&= b^A_{k,s} - b^D_{k,s},
\qquad s \in \mathcal{R}_k.
\label{eq:residual}
\end{align}
The overlap test allows the boxes to differ in extent while requiring spatial agreement.

For a frame $t$ between two consecutive correction frames $u,v\in\mathcal R_k$ with $u\leq t\leq v$, we linearly interpolate their residuals:
\begin{equation}
\tilde{r}_{k,t}
=
\frac{v-t}{v-u}r_{k,u}
+
\frac{t-u}{v-u}r_{k,v}.
\label{eq:residual_interp}
\end{equation}
Before the first correction frame and after the last, we use the nearest residual. If no correction frame is available, we retain the original visual track.

We add each interpolated or extended residual to $b^D_{k,t}$ and clip the corrected coordinates to $[0,1]$. If the resulting width or height is below $0.002$, we restore the original visual box. At correction frames, we use the amodal box directly. This step adjusts boxes at existing visual observation frames.

\paragraph{Gap interpolation.}
We retain the corrected visual box wherever a visual prediction exists and use the amodal box at frames without a visual prediction. At a shared frame, an amodal box that fails the overlap test neither defines a residual nor replaces the visual box.
Fusion includes only identities present in the amodal output.

Because both branches operate on sampled frames, the merged track may still contain missing frames. We fill these gaps by linearly interpolating box coordinates between consecutive available observations, provided that the endpoint boxes have an IoU of at least $\tau = 0.30$. Gaps that fail this test remain unfilled. We do not generate boxes before the first observation or after the last.

\begin{table}[t]
    \centering
    \caption{Comparison of Grounded VideoQA test HOTA on the Perception Test benchmark (932 videos and 1,859 questions). Baseline scores are taken from publicly reported results and cited references. SD-Track is evaluated on the same test set. All scores are scaled by 100.}
    \label{tab:perception_results}
    \small
    \setlength{\tabcolsep}{4pt}
    \begin{tabular}{lr}
        \toprule
        Method / team & Test HOTA $\uparrow$ \\
        \midrule
        MDETR+static~\cite{qtrace_perception2026} & 5.20 \\
        Wind Rain Tower, Qwen3.8-Max~\cite{qtrace_qwen_generalist2026} & 46.49 \\
        SGVR@KAIST~\cite{qtrace_sgvr2025} & 49.68 \\
        MR\_CAS~\cite{qtrace_mrcas2026} & 55.52 \\
        SRUKR~\cite{qtrace_srukr2026} & 65.22 \\
        Huawei SpaceMind~\cite{qtrace_huawei2026} & 65.41 \\
        \midrule
        SD-Track, dense SAM3 (ours) & 62.63 \\
        SD-Track, fusion (ours) & \textbf{66.20} \\
        \bottomrule
    \end{tabular}
\end{table}

\section{Experiments}
\label{sec:experiments}

\subsection{Setup}
\label{sec:experimental_setup}

\begin{figure*}[t]
    \centering
    \includegraphics[width=\textwidth]{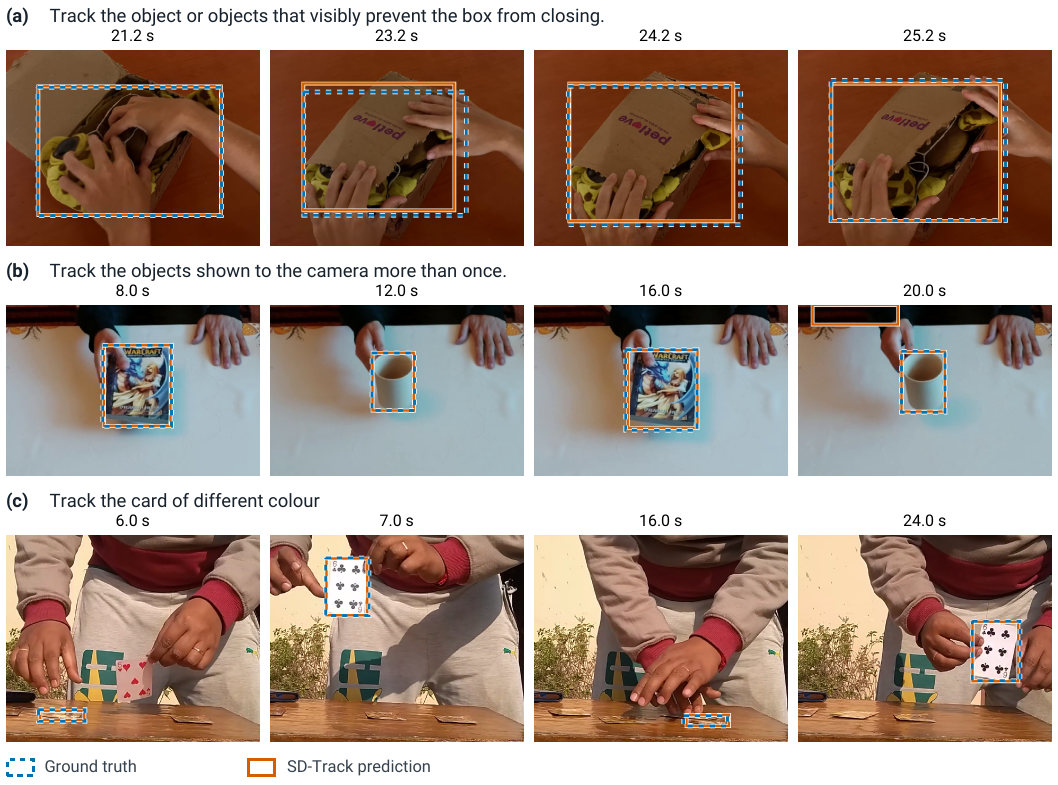}
    \caption{\textbf{Question-guided tracking examples from Perception Test Grounded VideoQA validation.} Each row shows a benchmark question and four video frames in temporal order. Blue dashed boxes denote ground truth, and orange solid boxes denote SD-Track  fusion predictions. The examples show (a) a plush turtle partially occluded by a cardboard box lid and hands, (b) a book and a cup shown to the camera more than once, and (c) a card of a different colour.}
    \label{fig:perception-task-examples}
\end{figure*}

We evaluate SD-Track on question-guided tracking, referring-expression grounding, and multi-instance action grounding. Perception Test Grounded VideoQA~\cite{Patraucean2023PerceptionTest} serves as our primary benchmark, requiring models to identify the objects that answer a question and recover their trajectories. Fig.~\ref{fig:perception-task-examples} illustrates benchmark questions and their corresponding ground-truth answer trajectories. We additionally evaluate declarative queries from VidSTG~\cite{Zhang2020VidSTG} for referring-expression grounding, and HC-STVG v1/v2~\cite{Tang2021HCSTVG}, both of which require localizing the person described by a query. We use SVAG-Bench~\cite{Hannan2025SVAG} to evaluate multi-instance action grounding.

We use Gemini 3.7 Flash for MLLM inference, except on VidSTG and HC-STVG, where we use Gemini 3.8 Flash. We fix the fusion threshold at $\tau=0.30$ (Sec.~\ref{sec:fusion}). For Perception Test, dense SAM3 tracking samples every third source frame, while sparse MLLM localization samples every thirtieth frame. These correspond to 10 fps and 1 fps for 30-fps videos. Each sparse call processes up to eight images.
Accuracy/cost and extent analyses use a 200-question subset of the Perception Test Validation set.
We provide detailed settings for the remaining experiments in the supplementary material.

\subsection{Evaluation Metrics}
\label{sec:evaluation_metrics}

We use Higher Order Tracking Accuracy (HOTA)~\cite{qtrace_hota} as the metric. At a box-matching IoU threshold $\alpha$, HOTA combines detection accuracy (DetA) and association accuracy (AssA). DetA measures the fraction of correctly matched detections relative to true positives, false positives, and false negatives. AssA averages association IoU over matched detections, penalizing temporal identity errors. The final score averages their geometric mean over localization thresholds:
\begin{equation}
    \operatorname{HOTA}=\frac{1}{|\mathcal{A}|}
    \sum_{\alpha\in\mathcal{A}}
    \sqrt{\operatorname{DetA}_{\alpha}\operatorname{AssA}_{\alpha}},
    \label{eq:hota}
\end{equation}
where $\mathcal{A}=\{0.05,0.10,\ldots,0.95\}$. Matching is recomputed at each threshold, and the geometric mean is taken before averaging over thresholds. The evaluation threshold $\alpha$ is distinct from the fusion threshold $\tau$.
We report question-wise mean HOTA on a 0 to 100 scale. Validation scores are computed on annotated frames with the local evaluator following the benchmark protocol.

For VidSTG and HC-STVG, we report mean temporal IoU (m\_tIoU) and mean video IoU (m\_vIoU). Temporal IoU measures overlap between predicted and ground-truth intervals. Video IoU sums frame-wise box IoU over their temporal intersection and divides by the number of frames in their temporal union. We report vIoU@0.3 and vIoU@0.5 as the percentages of queries exceeding each video IoU threshold.

For SVAG-Bench, we follow the proposed evaluation protocol and report HOTA for spatial grounding. For temporal grounding, we match predicted and ground-truth track identities using the HOTA matches at $\alpha=0.5$, then report mean temporal IoU and top-1 recall at a temporal IoU threshold of 0.5 (R1@0.5). The primary metric, m-HIoU, averages HOTA and mIoU within each split and then averages the three split scores.

\subsection{Results on Perception Test}
\label{sec:perception_results}

Table~\ref{tab:perception_results} compares SD-Track with baseline results reported in the official 2025 and 2026 Perception Test Grounded VideoQA leaderboards~\cite{qtrace_perception2026}. Baselines include static localization (MDETR+static), direct MLLM prediction (Wind Rain Tower~\cite{qtrace_qwen_generalist2026}), and pipelines combining semantic reasoning with visual grounding and tracking (SGVR@KAIST~\cite{qtrace_sgvr2025}, MR\_CAS~\cite{qtrace_mrcas2026}, SRUKR~\cite{qtrace_srukr2026}, and Huawei SpaceMind~\cite{qtrace_huawei2026}).

SD-Track achieves 66.20 HOTA, setting a new state of the art. Fusion improves our dense SAM3 result from 62.63 to 66.20, a gain of 3.57 points, showing that sparse MLLM localization complements dense tracking on this benchmark.
% Draft author note: SD-Track  rows currently use the same 2,325-query subset.
% Replace with full-split results and recheck protocol alignment before submission.
% The 2,423-query fusion run includes earlier results and is not used here.

\begin{table*}[t]
    \centering
    \caption{Comparison with supervised and zero-shot methods on VidSTG Declarative and HC-STVG v1/v2. SD-Track leads the compared zero-shot methods on all metrics and achieves the best overall scores on VidSTG Declarative and HC-STVG v1. Bold marks the best zero-shot score. Asterisks ($^{*}$) and underlining mark the best and second-best overall scores, respectively.}
    \label{tab:vidstg_declarative}
    \small
    \setlength{\tabcolsep}{4pt}
    \resizebox{\textwidth}{!}{%
    \begin{tabular}{@{}l*{12}{c}@{}}
        \toprule
        & \multicolumn{4}{c}{VidSTG Declarative}
        & \multicolumn{4}{c}{HC-STVG v1}
        & \multicolumn{4}{c}{HC-STVG v2} \\
        \cmidrule(lr){2-5}\cmidrule(lr){6-9}\cmidrule(lr){10-13}
        Method
        & m\_tIoU & m\_vIoU & \shortstack{vIoU\\@0.3} & \shortstack{vIoU\\@0.5}
        & m\_tIoU & m\_vIoU & \shortstack{vIoU\\@0.3} & \shortstack{vIoU\\@0.5}
        & m\_tIoU & m\_vIoU & \shortstack{vIoU\\@0.3} & \shortstack{vIoU\\@0.5} \\
        \midrule
        \multicolumn{13}{@{}l}{\textit{Supervised task-specific models}} \\
        TubeDETR~\cite{Yang2022TubeDETR}
        & 48.1 & 30.4 & 42.5 & 28.2
        & 43.7 & 32.4 & 49.8 & 23.5
        & 53.9 & 36.4 & 58.8 & 30.6 \\
        CG-STVG~\cite{Gu2024CGSTVG}
        & 51.4 & 34.0 & 47.7 & 33.1
        & 52.8 & 38.4 & 61.5 & 36.3
        & 60.0 & 39.5 & 64.5 & 36.3 \\
        TA-STVG~\cite{Gu2025TASTVG}
        & 51.7 & 34.4 & 48.2 & 33.5
        & 53.0 & 39.1 & 63.1 & 36.8
        & 60.4 & 40.2 & 65.8 & 36.7 \\
        ScanFocus~\cite{chen2026scanfocus}
        & 53.3 & 36.5 & 50.6 & 36.0
        & 55.5 & 41.8 & 67.5 & 42.2
        & 62.4 & 41.7 & 68.4 & 39.3 \\
        \midrule
        \multicolumn{13}{@{}l}{\textit{MLLMs trained for STVG}} \\
        SpaceVLLM~\cite{Wang2025SpaceVLLM}
        & 47.7 & 27.4 & 39.1 & 26.2
        & 56.9 & 39.3 & 66.6 & 36.9
        & 58.0 & 34.0 & 56.9 & 24.7 \\
        Bridge-STG~\cite{Tu2026BridgeSTG}
        & 52.6 & 37.2 & 52.4 & 37.4
        & -- & -- & -- & --
        & 64.1$^{*}$ & 41.5 & 67.5 & 38.6 \\
        STVG-o1~\cite{Gu2025STVGo1}
        & 52.1 & 33.5 & 48.4 & 32.0
        & \underline{60.3} & 44.1 & 73.3 & 43.5
        & \underline{63.8} & 41.2 & 68.5 & 39.6 \\
        Locate Anything~\cite{Rasheed2026LocateAnything}
        & \underline{53.7} & \underline{38.3} & \underline{53.0} & \underline{37.5}
        & 59.4 & \underline{45.6} & \underline{73.8} & \underline{46.7}
        & 63.1 & 43.6$^{*}$ & 71.0$^{*}$ & 42.2$^{*}$ \\
        \midrule
        \multicolumn{13}{@{}l}{\textit{Zero-shot methods}} \\
        LN-STVG~\cite{Yang2025LNSTVG}
        & -- & 18.0 & 29.8 & 12.2
        & -- & 24.8 & 41.5 & 16.3
        & -- & 27.7 & 44.7 & 19.5 \\
        RealVG~\cite{Wei2025RealVG}
        & 37.4 & 29.0 & 35.0 & 25.5
        & -- & 29.5 & 40.0 & 25.8
        & -- & 33.1 & 42.3 & 27.0 \\
        ASTG~\cite{zhao2026agentic}
        & 45.6 & 29.2 & 40.3 & 27.8
        & -- & 32.3 & 54.4 & 28.2
        & -- & 34.8 & 55.5 & 31.4 \\
        \midrule
        \multicolumn{13}{@{}l}{\textit{Ours (zero-shot)}} \\
        \textbf{SD-Track}
        & \textbf{55.62}$^{*}$ & \textbf{40.47}$^{*}$ & \textbf{57.61}$^{*}$ & \textbf{42.62}$^{*}$
        & \textbf{62.05}$^{*}$ & \textbf{47.39}$^{*}$ & \textbf{77.24}$^{*}$ & \textbf{51.47}$^{*}$
        & \textbf{63.65} & \underline{\textbf{43.19}} & \underline{\textbf{69.80}} & \underline{\textbf{42.00}} \\
        \bottomrule
    \end{tabular}}
\end{table*}

\begin{table}[t]
\centering
\caption{Comparison on the SVAG-Bench~\cite{Hannan2025SVAG}
test set (OVIS, MOT17, and MOT20).
We compare SD-Track with SVAGFormer (competition setting)
and LLaVA-Video~\cite{yang2025mot}.
HOTA measures spatial grounding, while mIoU and R1@0.5
evaluate temporal grounding.
m-HIoU is the mean of HOTA and mIoU averaged across
the three datasets. Higher is better.}
\label{tab:svagbench}
\small
\setlength{\tabcolsep}{3pt}
\resizebox{\columnwidth}{!}{%
\begin{tabular}{lcccc}
\toprule
Method
& HOTA$\uparrow$
& mIoU$\uparrow$
& R1@0.5$\uparrow$
& m-HIoU$\uparrow$ \\
\midrule

SVAGFormer~\cite{Hannan2025SVAG}
& 9.16 & 19.14 & 17.75 & 14.15 \\

LLaVA-Video~\cite{yang2025mot}
& 10.73 & 30.63 & 22.89 & 20.68 \\

\textbf{SD-Track  (ours)}
& \textbf{13.31}
& \textbf{36.28}
& \textbf{36.45}
& \textbf{24.80} \\

\bottomrule
\end{tabular}%
}
\end{table}

% \begin{table}[t]
% \centering
% \caption{Comparison on the SVAG-Bench~\cite{Hannan2025SVAG}
% test set (OVIS, MOT17, and MOT20).
% HOTA, DetA, and AssA evaluate spatial grounding,
% while mIoU and R1@0.5 evaluate temporal grounding.
% m-HIoU averages HOTA and mIoU across the three datasets.
% Higher is better.}
% \label{tab:svagbench}
% \small
% \setlength{\tabcolsep}{3pt}
% \resizebox{\columnwidth}{!}{%
% \begin{tabular}{lcccccc}
% \toprule
% Method
% & HOTA$\uparrow$
% & DetA$\uparrow$
% & AssA$\uparrow$
% & mIoU$\uparrow$
% & R1@0.5$\uparrow$
% & m-HIoU$\uparrow$ \\
% \midrule

% SVAGFormer~\cite{Hannan2025SVAG}
% & 9.16 & 4.09 & 27.70 & 19.14 & 17.75 & 14.15 \\

% LLaVA-Video~\cite{yang2025mot}
% & 10.73 & 4.07 & \textbf{41.69}
% & 30.63 & 22.89 & 20.68 \\

% \textbf{SD-Track (ours)}
% & \textbf{13.31}
% & \textbf{6.88}
% & 35.00
% & \textbf{36.28}
% & \textbf{36.45}
% & \textbf{24.80} \\

% \bottomrule
% \end{tabular}%
% }
% \end{table}

\subsection{Results on Additional Benchmarks}
\label{sec:additional_benchmarks}

\paragraph{VidSTG and HC-STVG.}
VidSTG~\cite{Zhang2020VidSTG} evaluates spatial and temporal localization of referred objects in untrimmed videos. We evaluate its declarative queries, which describe targets through their appearance, actions, or relationships. Following prior work~\cite{Gu2024CGSTVG, Wang2025SpaceVLLM}, we report results on the HC-STVG v1 test set and the HC-STVG v2 validation set.

The comparison includes supervised task-specific models, supervised MLLMs, and zero-shot methods. Supervised task-specific models, such as TubeDETR~\cite{Yang2022TubeDETR}, CG-STVG~\cite{Gu2024CGSTVG}, and TA-STVG~\cite{Gu2025TASTVG}, learn to predict temporal boundaries and box trajectories. MLLMs trained for STVG, including SpaceVLLM~\cite{Wang2025SpaceVLLM} and Bridge-STG~\cite{Tu2026BridgeSTG}, connect multimodal representations to learned spatial decoders. The zero-shot baselines use attention cues (LN-STVG~\cite{Yang2025LNSTVG}), query decomposition (RealVG~\cite{Wei2025RealVG}), or collaborative spatial and temporal reasoning (ASTG~\cite{zhao2026agentic}). These methods generally underperform the supervised task-specific models. For example, their m\_vIoU scores are lower across all three benchmarks in Table~\ref{tab:vidstg_declarative}.

SD-Track combines sparse MLLM amodal localization with dense tracking through a shared plan, without benchmark-specific training. As reported in Table~\ref{tab:vidstg_declarative}, it leads the compared zero-shot methods on all three benchmarks and achieves the best overall scores across all four metrics on VidSTG Declarative and HC-STVG v1. Its m\_vIoU exceeds ASTG, the strongest compared zero-shot baseline, by 11.27, 15.09, and 8.39 points on VidSTG Declarative and HC-STVG v1/v2, respectively.

\paragraph{SVAG.}
We further evaluate SD-Track on SVAG-Bench~\cite{Hannan2025SVAG},
which requires identifying and tracking all object instances
matching a natural-language action description while localizing
their corresponding temporal intervals.
As shown in Table~\ref{tab:svagbench}, we compare against
SVAGFormer~\cite{Hannan2025SVAG} and the training-free
LLaVA-Video-based approach~\cite{yang2025mot}
on the official test set comprising OVIS, MOT17, and MOT20.
SD-Track achieves an m-HIoU of 24.80, outperforming
SVAGFormer (14.15) and LLaVA-Video (20.68).
It also achieves the highest HOTA (13.31) and temporal
mIoU (36.28), demonstrating strong spatial and temporal
grounding performance across the benchmark.

\subsection{Ablation Study}
\subsubsection{Accuracy and API Cost}
\label{sec:accuracy_cost}

We compare dense SAM3 tracking, two Gemini-only baselines, and SD-Track on a 200-question subset of the Perception Test validation set. Sparse Gemini reuses the same amodal predictions sampled at 1 fps that are used by SD-Track, then linearly interpolates box coordinates between consecutive observations without using SAM3. Direct Gemini obtains amodal predictions covering each object's full extent, including occluded regions, at every original video frame (approximately 30 fps). Both Gemini baselines therefore perform amodal localization, with inference at 1 fps and approximately 30 fps, respectively. All output trajectories are indexed on the original video frame grid. SD-Track combines the same 1-fps Gemini amodal predictions with dense SAM3 tracks. It follows the dense trajectories to maintain temporal instance correspondence and interpolates residual corrections from spatially consistent sparse amodal observations along those tracks.

\begin{table}[t]
    \centering
    \caption{Accuracy and estimated Gemini API cost~\cite{qtrace_gemini_pricing} on a 200-question subset of the Perception Test validation set. Costs include planning for all configurations and amodal localization where used. SAM3 GPU compute is excluded.}
    \label{tab:system_comparison}
    \small
    \setlength{\tabcolsep}{4pt}
    \begin{tabular}{lrr}
        \toprule
        Configuration & HOTA $\uparrow$ & API cost (USD) $\downarrow$ \\
        \midrule
        Dense SAM3 (visual) & 66.86 & 7.54 \\
        Sparse Gemini (amodal) & 65.61 & 20.91 \\
        Direct Gemini (amodal) & \textbf{72.55} & 206.18 \\
        SD-Track (fusion) & 71.80 & 20.91 \\
        \bottomrule
    \end{tabular}
\end{table}

Direct Gemini achieves the highest HOTA of 72.55 (Table~\ref{tab:system_comparison}). SD-Track scores 71.80, improving on dense SAM3 by 4.94 points and falling 0.75 points below direct Gemini. Using the same planning and sparse localization calls, it outperforms sparse Gemini with linear interpolation (65.61) by 6.19 points.

SD-Track and sparse Gemini each have an API cost of 20.91 USD, 89.9\% less than the 206.18 USD for direct Gemini. Dense SAM3 shares SD-Track's plan and dense predictions, with an API cost of 7.54 USD for planning alone. Their shared SAM3 tracking and model loading take 2.31 GPU-hours on the 200-question subset. This GPU compute is not needed by the sparse Gemini baseline and is excluded from the API cost comparison.

\subsubsection{Amodal Extent Correction}
\label{sec:amodal_extent}
\label{sec:qualitative_amodal}

SAM3 boxes can underestimate the target's extent under partial occlusion. To measure the benefit of expanding these boxes using amodal predictions, we compare full fusion with a variant that removes increases in width or height from the fusion corrections. This variant retains shifts in box centers and decreases in width or height, and still uses amodal boxes at frames without visual predictions. In Table~\ref{tab:extent_control}, disabling these expansion corrections reduces HOTA from 71.80 to 69.17, a drop of 2.63 points on the same 200-question validation subset.

Figure~4 illustrates how SD-Track transfers sparse amodal
corrections along dense visual trajectories for a moving
remote control partially occluded by fingers. Gemini provides amodal boxes at frames 300 and 330. At these reference frames, we compute coordinate residuals between the sparse amodal and dense visual boxes. We then interpolate the residuals and add them to existing SAM3 boxes at intermediate frames, including frames 306, 315, and 324, where Gemini is not queried. At frame 315, for example, the correction gives equal weight to the two reference residuals. This transfers the estimated hidden extent to frames between sparse observations while following the dense visual trajectory.

At the annotated reference frames, fusion improves box IoU from 0.384 to 0.869 at frame 300 and from 0.245 to 0.655 at frame 330. Across all annotated frames for this question, fusion increases HOTA from 65.28 to 83.09, a gain of 17.81 points.

\begin{table}[t]
    \centering
    \caption{Effect of box expansion during fusion on the 200-question Perception Test validation subset. The ablated variant disables increases in box width or height while retaining center shifts, size reductions, and amodal boxes at frames without visual predictions. Output frames and interpolation endpoint frames are fixed to those of full fusion.}
    \label{tab:extent_control}
    \small
    \setlength{\tabcolsep}{5pt}
    \begin{tabular}{lrr}
        \toprule
        Configuration & HOTA $\uparrow$ & $\Delta$ \\
        \midrule
        SD-Track  fusion & \textbf{71.80} & 0.00 \\
        Without box expansion & 69.17 & $-2.63$ \\
        \bottomrule
    \end{tabular}
\end{table}

\begin{figure}[t]
    \centering
    \includegraphics[width=\linewidth]{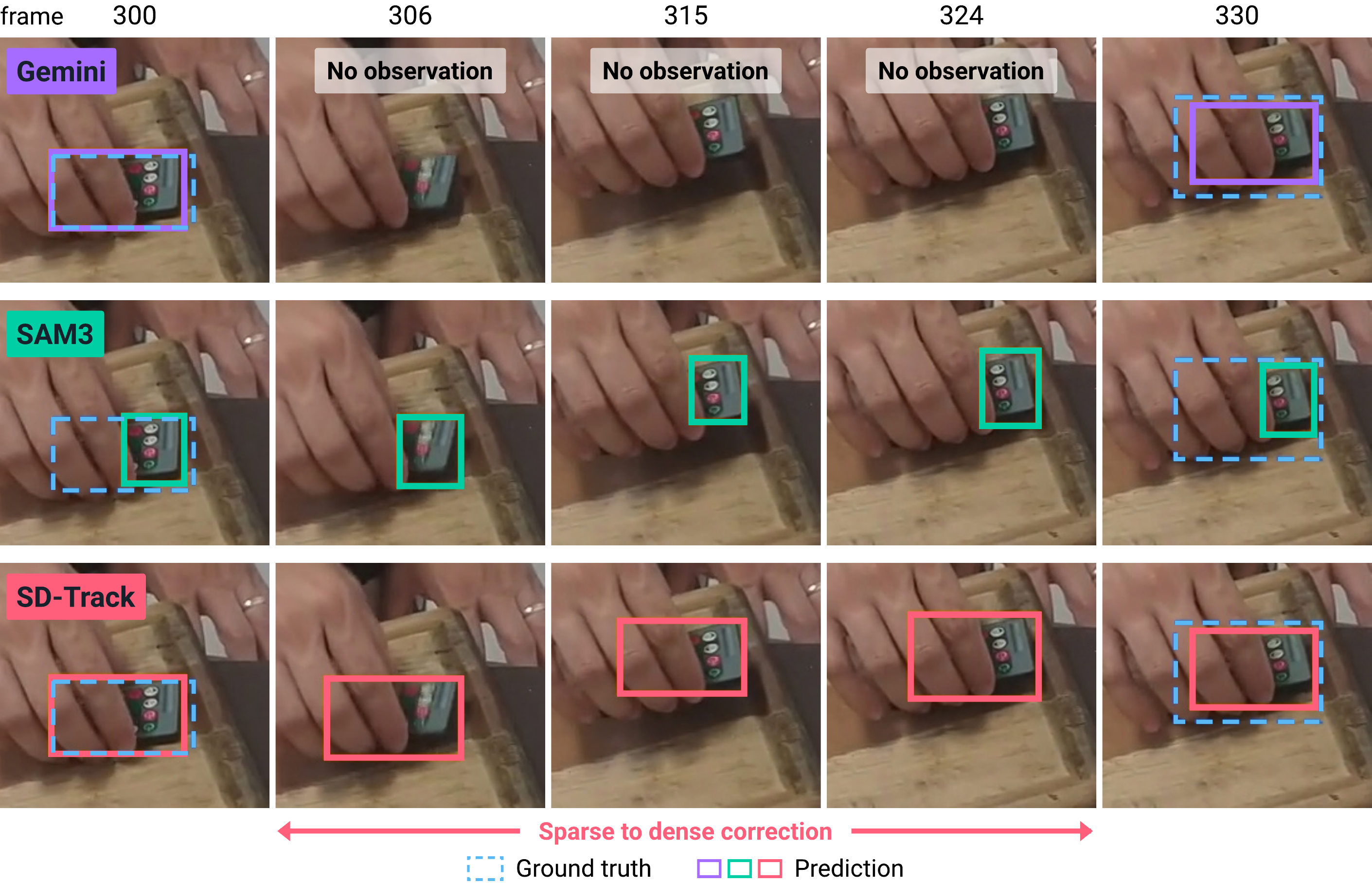}
    \caption{Sparse-to-dense amodal correction for a remote
    control partially occluded by fingers
    (Perception Test validation).
    Rows show Gemini, SAM3, and SD-Track predictions.
    Gemini provides sparse amodal boxes at frames 300 and 330,
    while SD-Track propagates their coordinate corrections
    along the dense SAM3 trajectory.
    Solid boxes denote predictions, and blue dashed boxes
    denote ground truth at annotated frames.}

    \label{fig:remote_residual}
\end{figure}

\section{Conclusion}
\label{sec:conclusion}

We introduced SD-Track, a zero-shot agentic framework that coordinates sparse MLLM amodal localization and dense visual tracking through a shared question-guided plan. SD-Track follows dense visual trajectories to maintain temporal instance correspondence and propagates residual corrections from spatially consistent sparse amodal observations to recover object extent under occlusion. This combination allows the two branches to compensate for errors at different points in the trajectory. SD-Track achieves state-of-the-art HOTA scores of 66.20 on Perception Test Grounded VideoQA and 13.31 on SVAG-Bench, and improves referring-expression grounding on VidSTG. Our validation analysis further shows that sparse amodal fusion approaches the accuracy of per-frame MLLM amodal localization at substantially lower API cost.

\section*{LLM Usage Statement}
Codex (OpenAI) assisted with drafting and revising manuscript text, organizing technical explanations, checking bibliographic references, and formatting LaTeX equations, algorithms, and tables. Gemini 3.7 Flash/3.8 Flash are used as a component of SD-Track for target identification, critic verification, shared tracking-plan construction, and sparse amodal localization, as described in the Methodology section. The authors retain full responsibility for the scientific claims, experimental results, references, and final manuscript.

% One shared bibliography, including citations made in the supplementary body.
{
    \small
    \bibliographystyle{ieeenat_fullname}
    \bibliography{main}
}

\clearpage
% Supplementary material included by main.tex.
% Do not compile this file on its own.

\graphicspath{{sec/figures/suppl_fig/}{sec/figures/}}

% Start supplementary numbering without resetting page numbers.
\setcounter{section}{0}
\setcounter{subsection}{0}
\setcounter{figure}{0}
\setcounter{subfigure}{0}
\setcounter{table}{0}
\setcounter{equation}{0}

\renewcommand{\thesection}{\Alph{section}}
\renewcommand{\thefigure}{S\arabic{figure}}
\renewcommand{\thetable}{S\arabic{table}}
\renewcommand{\theequation}{S\arabic{equation}}

% Keep supplementary hyperlink targets distinct from main-paper targets.
\renewcommand{\theHsection}{supp.\arabic{section}}
\renewcommand{\theHsubsection}{\theHsection.\arabic{subsection}}
\renewcommand{\theHfigure}{supp.\arabic{figure}}
\renewcommand{\theHtable}{supp.\arabic{table}}
\renewcommand{\theHequation}{supp.\arabic{equation}}
\providecommand{\theHsubfigure}{}
\renewcommand{\theHsubfigure}{\theHfigure.\arabic{subfigure}}

\microtypesetup{protrusion=true,expansion=true}

% Allow the two occlusion tables and failure example to share a page.
\setcounter{dbltopnumber}{3}
\renewcommand{\dbltopfraction}{0.90}
\renewcommand{\dblfloatpagefraction}{0.55}
\setlength{\dbltextfloatsep}{10pt plus 2pt minus 2pt}
\setlength{\dblfloatsep}{9pt plus 2pt minus 2pt}
\setlength{\emergencystretch}{1.25em}

\raggedbottom

% PAGE 1: title and introductory text in two columns.
\twocolumn[{%
\begin{minipage}{\textwidth}
\centering
{\Large\bfseries\SDTrackTitle\par}
\vspace{0.5em}
{\Large Supplementary Material\par}
\medskip
\end{minipage}
}]

\section{Additional Qualitative Comparisons}
\label{sec:supp_qualitative}

We present six additional qualitative examples to examine the
complementary behavior of SAM3 and Gemini.
Figures~\ref{fig:supp_purple}--\ref{fig:supp_gray} compare the three
methods. Rows show SAM3 (blue), Gemini (orange), and SD-Track (green).
Dashed white boxes indicate ground truth, and solid colored boxes
denote predictions.

Under occlusion, SAM3 can produce boxes that cover only part of
the annotated target, whereas Gemini can provide more complete
target coverage but may confuse visually similar instances.
Yellow crosses in the upper-left corners of the frames indicate
incomplete target coverage under occlusion, while red crosses
indicate object confusion.

SD-Track retains the intended target identity while improving
target coverage in the highlighted occlusion cases.
These qualitative observations illustrate the complementary
strengths of dense temporal tracking and sparse semantic guidance.

\section{Benchmark and Implementation Details}
\label{sec:supp_benchmarks}

We evaluate SD-Track on four benchmark families covering question-guided
tracking, single-object spatio-temporal grounding, and multi-instance action
grounding, without benchmark-specific training.
Table~\ref{tab:supp_benchmark_overview} reports the actual evaluation cohorts,
inference schedules, and evaluation grids. HC-STVG v1 and v2 are listed
separately because they use different evaluation partitions.

\subsection{Perception Test Grounded VideoQA}
\label{sec:supp_perception}

Perception Test~\cite{Patraucean2023PerceptionTest} requires identifying
the objects that answer a question and predicting their box trajectories.
We evaluate the full test partition of 1,859 questions over 932 videos
and use a fixed 200-question validation subset for development analyses.
The full validation partition contains 3,051 questions over 1,545
videos~\cite{Heyward2026PerceptionTest2025}.
Boxes are annotated at approximately 1\,fps, including approximate
locations during occlusion. HOTA~\cite{qtrace_hota} is evaluated on the
actual annotated frame indices and averaged over questions; GT boxes are
not interpolated. This annotation cadence differs from inference sampling.
The task permits using later observations to resolve earlier identities.

\subsection{VidSTG}
\label{sec:supp_vidstg}

VidSTG~\cite{Zhang2020VidSTG} pairs referring descriptions with a target
object's temporal interval and box trajectory.
We evaluate all 4,610 declarative test queries, covering 2,527 input clips;
multiple queries can share a clip. Dense and sparse inference use
source-clock rates of 10\,fps and 1\,fps, respectively.
Evaluation follows the TubeDETR-compatible 5-fps grid with at most 200
frames~\cite{Yang2022TubeDETR}; this cap applies to evaluation, not inference.

\subsection{HC-STVG v1 and v2}
\label{sec:supp_hcstvg}

HC-STVG~\cite{Tang2021HCSTVG} localizes a described person in a 20\,s
clip using appearance, actions, and interactions.
Following TubeDETR~\cite{Yang2022TubeDETR}, we evaluate all 1,160 v1 test
pairs and all 2,000 v2 validation pairs, with one query per input clip.
Evaluation uses the native source-frame grid.
Dense and sparse inference sample every third and thirtieth source frame.
For the observed 15--30-fps clips, these correspond to 5--10\,fps and
0.5--1\,fps rather than fixed 10\,fps and 1\,fps rates.

\subsection{SVAG-Bench}
\label{sec:supp_svag}

SVAG-Bench~\cite{Hannan2025SVAG} requires recovering every instance
matching an action description and its temporal interval.
Our test cohort contains 902 OVIS queries from 137 videos, 499 MOT17
queries from seven videos, and 432 MOT20 queries from two videos.
These are unique evaluated queries, distinct from the published totals
of annotation records containing language variants and instance references.
We follow the benchmark's source partitions: its OVIS test subset comes
from the original OVIS validation split, while its MOT20 partitions divide
the original MOT20 training split.
The dense base grid targets $\min(f_v,10)$\,fps, so all frames of the
5-fps OVIS videos are used. Sparse localization targets 1\,fps on the
original video clock.

\subsection{Implementation and Reproducibility}
\label{sec:supp_implementation}

\paragraph{Models and input preparation.}
We use \texttt{gemini-3.7-flash} for planning and sparse localization on
Perception Test and SVAG-Bench, and \texttt{gemini-3.8-flash} on VidSTG
and HC-STVG. Planning uses a uniformly sampled proxy of at most 192
source frames, labeled with their original indices and resized with
aspect ratio preserved to a maximum dimension of 512 pixels.
The proxy is encoded at 6\,fps and supplied with Gemini video-sampling
metadata of 4\,fps; these proxy-clock settings are separate from the
source-clock anchor grids in Table~\ref{tab:supp_benchmark_overview}.
Selected anchors are localized on the exact source images.
Sparse localization receives native-resolution JPEGs without resizing,
in groups of up to eight. For SVAG-Bench, the group size is
$\min(8,\max(1,\lfloor64/K\rfloor))$ for $K$ saved target identities.
Requests use temperature 0, schema-constrained outputs, and high thinking
for sparse localization. Per-run generation settings and full prompt
schemas accompany the code.
Dense tracking uses the pretrained SAM3 \texttt{sam3.pt} checkpoint and
its standard $1008\!\times\!1008$ internal input configuration.

% PAGES 2--3: Figure S1 is placed on two dedicated full-width pages.
% \onecolumn ends page 1; do not add another page break before it.
\onecolumn

\noindent\begin{minipage}[t][0.97\textheight][s]{\textwidth}
\centering
\captionsetup{type=figure}
    \begin{subfigure}{0.90\textwidth}
        \centering
        \includegraphics[width=\linewidth]{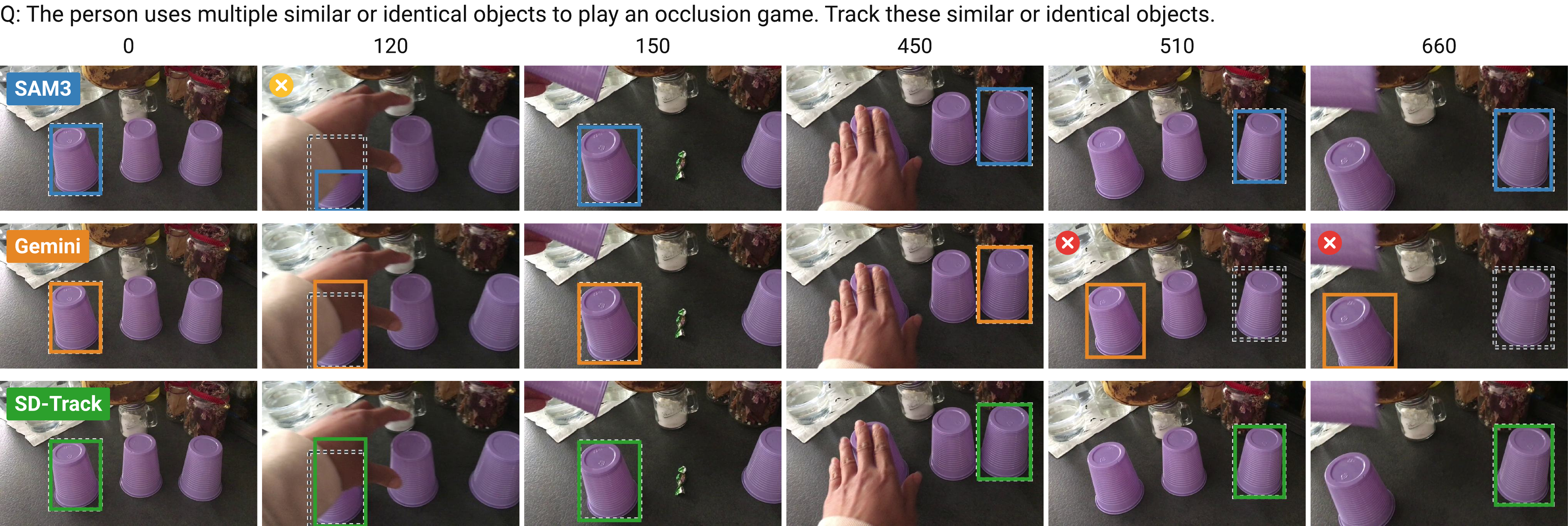}
        \caption{\textbf{Purple-cup example.} Comparison of the dense tracker, sparse MLLM localization, and their fusion.}
        \label{fig:supp_purple}
    \end{subfigure}
    \par\vfill
    \begin{subfigure}{0.90\textwidth}
        \centering
        \includegraphics[width=\linewidth]{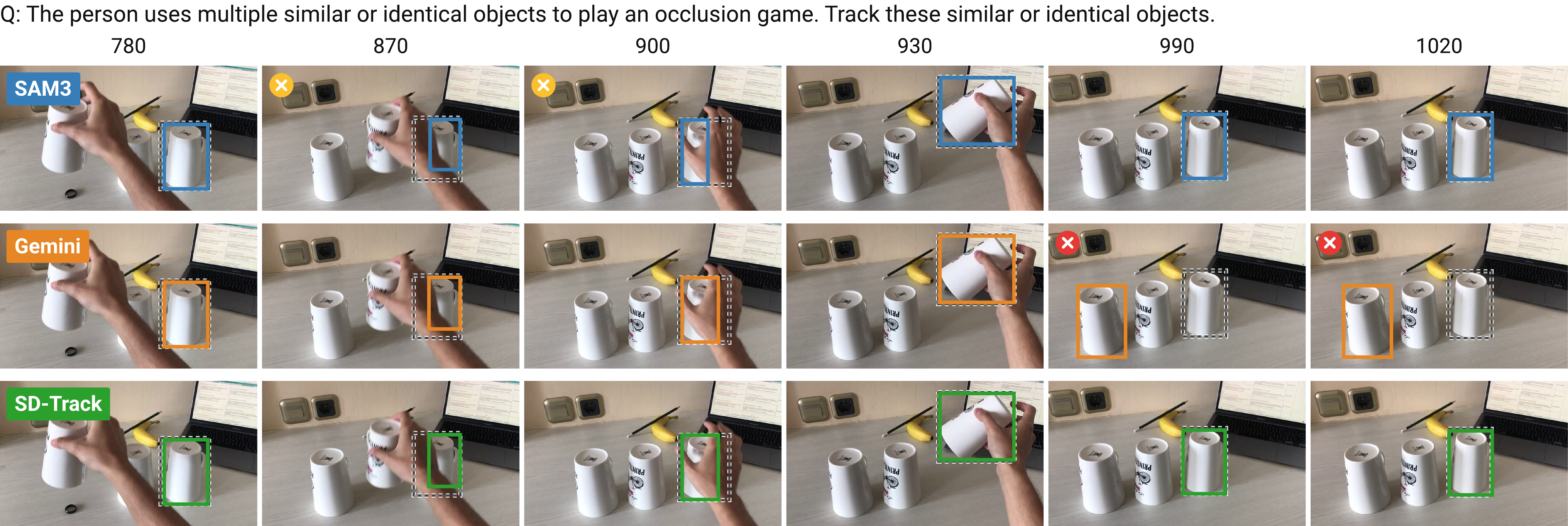}
        \caption{\textbf{White-cup example.} Occlusion may reduce visible-mask coverage while the amodal branch estimates a larger extent.}
        \label{fig:supp_white}
    \end{subfigure}
    \par\vfill
    \begin{subfigure}{0.90\textwidth}
        \centering
        \includegraphics[width=\linewidth]{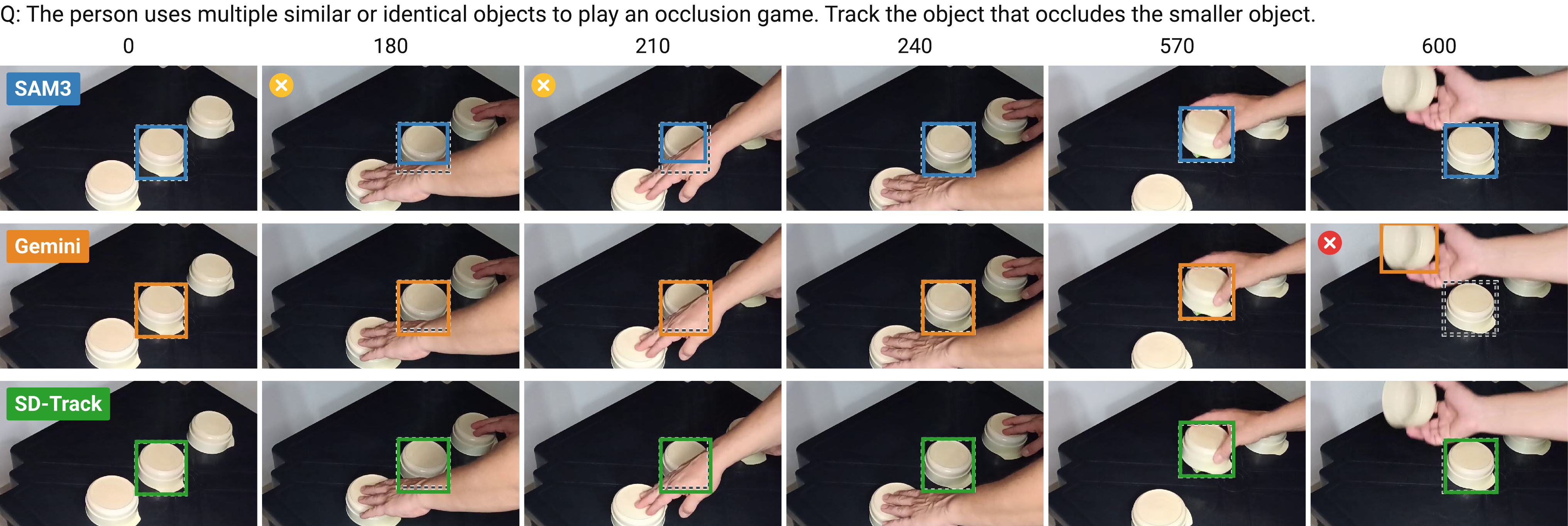}
        \caption{\textbf{Bowl example.} Additional qualitative comparison among SAM3, Gemini, and SD-Track.}
        \label{fig:supp_bowl}
    \end{subfigure}
    \par\smallskip
    \caption{\textbf{Additional qualitative comparisons among SAM3, Gemini, and SD-Track.}
    Examples are shown in source order: (a) Purple cup, (b) White cup, and (c) Bowl.
    Dashed white boxes denote ground truth, and colored boxes denote predictions.
    Yellow \texttimes\ marks occlusion frames, and red \texttimes\ marks object-confusion frames.}
    \label{fig:supp_qualitative}
\end{minipage}\par

\newpage

\noindent\begin{minipage}[t][0.97\textheight][s]{\textwidth}
\centering
\captionsetup{type=figure}
    \ContinuedFloat
    \centering

    \begin{subfigure}{0.90\textwidth}
        \centering
        \includegraphics[width=\linewidth]{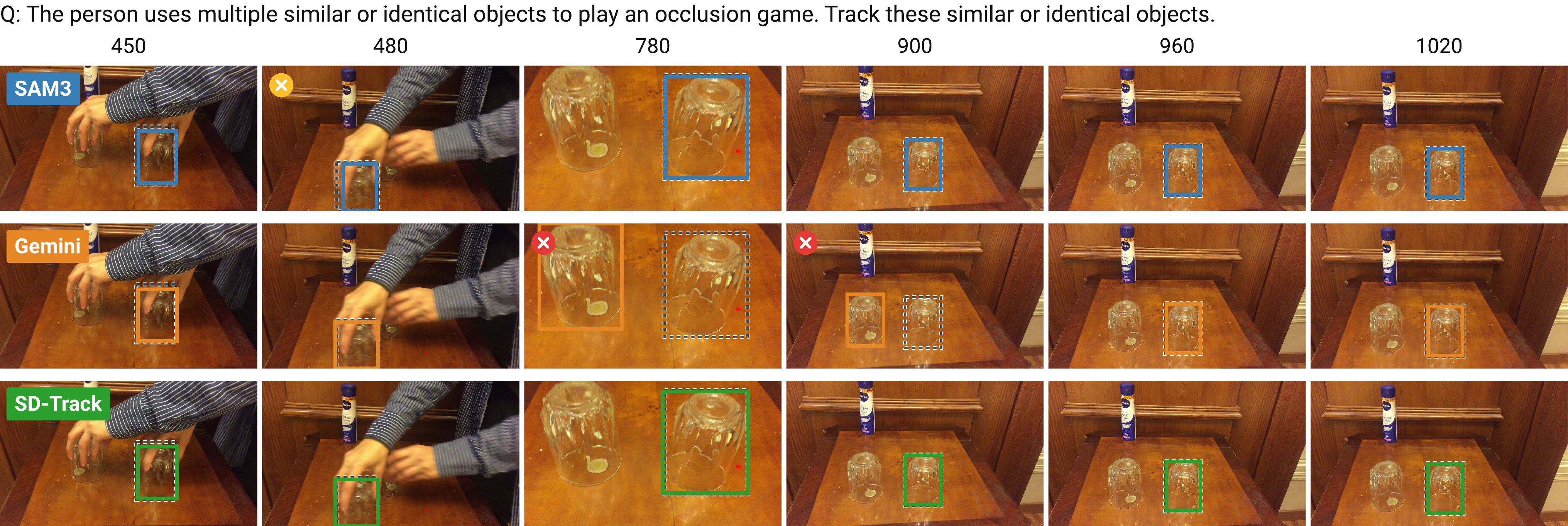}
        \caption{\textbf{Transparent-glass example.} Similar objects can produce identity confusion in sparse localization.}
        \label{fig:supp_glass}
    \end{subfigure}

    \par\vfill

    \begin{subfigure}{0.90\textwidth}
        \centering
        \includegraphics[width=\linewidth]{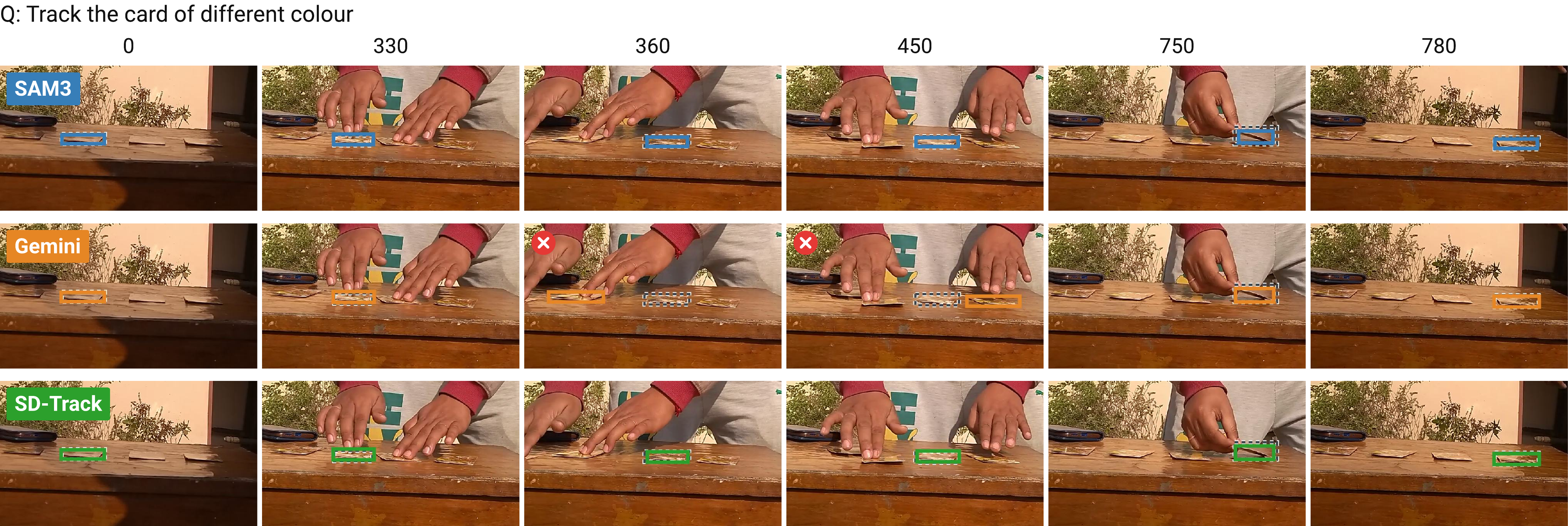}
        \caption{\textbf{Card example.} Target selection is tested during appearance changes and interactions.}
        \label{fig:supp_card}
    \end{subfigure}

    \par\vfill

    \begin{subfigure}{0.90\textwidth}
        \centering
        \includegraphics[width=\linewidth]{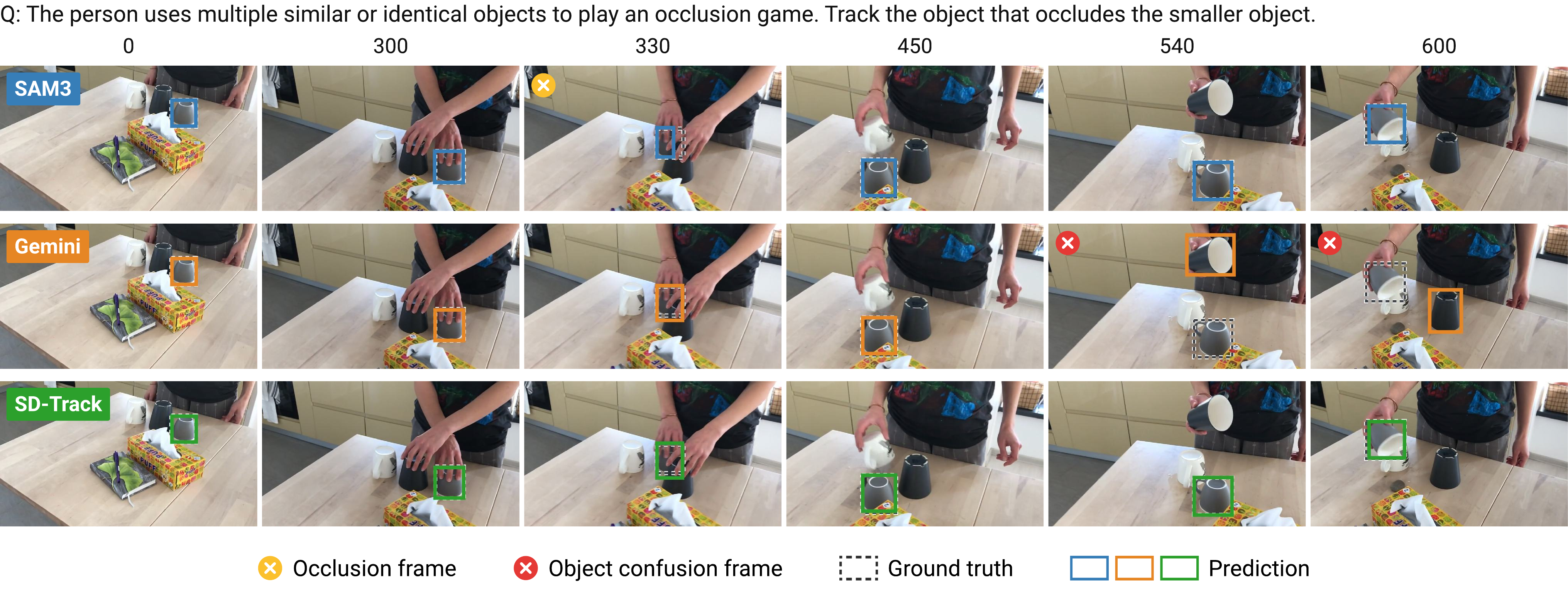}
        \caption{\textbf{Small-gray-cup example.} SD-Track combines dense correspondence with sparse spatial corrections.}
        \label{fig:supp_gray}
    \end{subfigure}

    \caption[]{\textbf{Additional qualitative comparisons among SAM3, Gemini, and SD-Track (continued).}
    Examples are shown in source order: (d) Transparent glass, (e) Card, and (f) Small gray cup.
    Dashed white boxes denote ground truth, and colored boxes denote predictions.
    Yellow \texttimes\ marks occlusion frames, and red \texttimes\ marks object-confusion frames.}
\end{minipage}\par

% PAGE 4 ONWARD: resume the original two-column body.
% \twocolumn itself ends the preceding figure page.
\twocolumn

\paragraph{Source frames and tracking.}
Perception Test uses source-frame strides of 10, 3, and 30 for anchor
candidates, dense inference, and sparse localization, respectively.
VidSTG dense and sparse grids use timestamp sampling
$f_k = \lfloor k f_v/h + 0.5 \rfloor$ for target rate $h$.
The dense grid additionally includes the saved sparse indices.
All selected identities share a SAM3 session. Dense sampling additionally
includes selected anchors and frames within $\pm15$ source frames of the
planned visibility boundaries, with a visibility-gating margin of 15
source frames, on Perception Test, VidSTG, and HC-STVG.
SVAG-Bench uses $\lceil0.5f_v\rceil$ frames for both settings.
Visibility gating filters each identity's predictions after propagation;
it does not restrict the SAM3 propagation session to the planned interval.

\paragraph{Temporal output and coordinates.}
For VidSTG and HC-STVG, the planner predicts a
\texttt{grounding\_segment}, which is used to crop the final tube.
The exporter applies the saved clip-to-source frame offset.
For SVAG-Bench, per-identity \texttt{action\_segments} are predicted
separately from \texttt{visibility\_segments}. Inclusive intervals
$[a,b]$ are exported as half-open intervals $[a,b+1)$; spatial boxes
are serialized separately. GT intervals are not substituted for predicted
intervals. MLLM boxes use the $[y_{\min},x_{\min},y_{\max},x_{\max}]$
order on a 0--1000 scale and are converted to normalized corner coordinates
before fusion and to the benchmark's required format at export.

\paragraph{Validation subset and reproducibility.}
We construct val200 with seed 0 using six strata defined by annotated
answer-track count (one versus two or more) and question area (physics,
memory, or abstraction). Samples are allocated proportionally with
largest-remainder rounding and drawn within lexicographically ordered
strata using \texttt{random.Random(0)}.
The resulting 200 queries span 190 videos: 120 are single-target and
80 multi-target; 174 concern physics, 17 memory, and nine abstraction.
Answer-count metadata is used only to form the subset and is not supplied
to the runtime predictors.
This fixed cohort supports accuracy--cost comparisons, extent ablations,
threshold exploration, observation-holdout diagnostics, and occlusion
analysis; it is a development-validation cohort.
The reproduction materials provide the exact ordered query list,
selection code, configurations, prompts, frame maps, evaluator sources,
and source-file hashes. The retained SVAGEval version is
\texttt{65b1495}. Archived source hashes identify executed versions
independently of subsequent working-tree changes.

\begin{table*}[!t]
    \centering
    \caption{\textbf{Evaluation cohorts and sampling schedules.}
    Counts refer to the queries and input videos or clips evaluated by
    SD-Track. A, D, and S denote the anchor-candidate, dense-base-grid, and
    sparse sampling rates in fps; $f_v$ is the source video frame rate.
    Dense sampling additionally includes selected anchors and visibility-boundary
    frames. Perception Test rates are nominal values for 30-fps videos.
    All metrics are higher-is-better and reported on a 0--100 scale.}
    \label{tab:supp_benchmark_overview}
    \small
    \setlength{\tabcolsep}{3.6pt}
    \renewcommand{\arraystretch}{1.18}
    \begin{tabularx}{\textwidth}{@{}
        >{\raggedright\arraybackslash}p{0.14\textwidth}
        >{\raggedright\arraybackslash}p{0.21\textwidth}
        >{\centering\arraybackslash}p{0.065\textwidth}
        >{\centering\arraybackslash}p{0.085\textwidth}
        >{\centering\arraybackslash}p{0.065\textwidth}
        >{\raggedright\arraybackslash}p{0.14\textwidth}
        >{\raggedright\arraybackslash}X@{}}
        \toprule
        \textbf{Benchmark} & \textbf{Evaluation cohort} &
        \textbf{A (fps)} & \textbf{D (fps)} & \textbf{S (fps)} &
        \textbf{Evaluation grid} & \textbf{Metrics} \\
        \midrule
        Perception Test~\cite{Patraucean2023PerceptionTest} &
        Test: 1,859 queries; 932 videos &
        3 & 10 & 1 &
        Official test server & HOTA \\
        \addlinespace[2pt]
        Perception Test val200 &
        Validation: 200 queries; 190 videos &
        3 & 10 & 1 &
        Annotated GT frames & HOTA \\
        \addlinespace[2pt]
        VidSTG~\cite{Zhang2020VidSTG} &
        Declarative test: 4,610 queries; 2,527 clips &
        $f_v/10$ & 10 & 1 &
        5 fps; at most 200 frames &
        m\_tIoU, m\_vIoU, vIoU@0.3/0.5 \\
        \addlinespace[2pt]
        HC-STVG v1~\cite{Tang2021HCSTVG} &
        Test: 1,160 video--sentence pairs &
        $f_v/10$ & $f_v/3$ & $f_v/30$ &
        Native source frames &
        m\_tIoU, m\_vIoU, vIoU@0.3/0.5 \\
        \addlinespace[2pt]
        HC-STVG v2~\cite{Tang2021HCSTVG,Yang2022TubeDETR} &
        Validation: 2,000 video--sentence pairs &
        $f_v/10$ & $f_v/3$ & $f_v/30$ &
        Native source frames &
        m\_tIoU, m\_vIoU, vIoU@0.3/0.5 \\
        \addlinespace[2pt]
        SVAG-Bench~\cite{Hannan2025SVAG} &
        Test: 1,833 queries; 146 videos &
        $3^{\dagger}$ & $\min(f_v,10)$ & $1^{\ddagger}$ &
        Official SVAGEval protocol &
        HOTA, temporal mIoU, R1@0.5, m-HIoU \\
        \bottomrule
    \end{tabularx}
    \vspace{2pt}
    \begin{minipage}{\textwidth}\footnotesize
    $^{\dagger}$All frames are anchor candidates when $f_v\leq5$.
    $^{\ddagger}$A saved anchor can be added when a target has no visible
    observation on the 1-fps sparse grid.
    A describes the candidate grid, not the number of selected anchors.
    \end{minipage}
\end{table*}

\subsection{Metric and Annotation Conventions}
\label{sec:supp_benchmark_metrics}

\paragraph{Single-object spatio-temporal grounding.}
For VidSTG and HC-STVG, temporal IoU measures interval overlap, while
video IoU also accounts for spatial localization:
\begin{equation}
    \mathrm{vIoU}
    =
    \frac{\displaystyle\sum_{t\in T_p\cap T_g}
        \mathrm{IoU}\!\left(b_t^p,b_t^g\right)}
    {|T_p\cup T_g|},
    \label{eq:supp_viou}
\end{equation}
where $T_p$ and $T_g$ are the evaluation-frame grids covered by the
predicted and GT temporal intervals. A missing predicted box contributes
zero spatial IoU. We report query-wise m\_tIoU and m\_vIoU, and
vIoU@0.3/0.5, the percentages of queries whose vIoU exceeds the respective
threshold~\cite{Zhang2020VidSTG,Yang2022TubeDETR}.

\paragraph{Multi-instance grounding.}
For SVAG-Bench, HOTA measures detection and identity association;
mIoU and R1@0.5 measure temporal localization.
The official evaluator uses HOTA matches at spatial IoU $\alpha=0.5$
and majority voting across frames to associate predicted and reference
identities before temporal evaluation~\cite{Hannan2025SVAG}.
R1@0.5 tests whether the highest-ranked temporal prediction has
temporal IoU $\geq0.5$. The aggregate score is
\begin{equation}
    \mathrm{m\mbox{-}HIoU}
    =
    \frac{1}{|\mathcal{D}|}
    \sum_{d\in\mathcal{D}}
    \frac{\mathrm{HOTA}_{d}+\mathrm{mIoU}_{d}}{2},
    \label{eq:supp_mhiou}
\end{equation}
where $\mathcal{D}=\{\mathrm{OVIS},\mathrm{MOT17},\mathrm{MOT20}\}$
and mIoU denotes temporal overlap. Each source receives equal weight,
regardless of its number of videos or queries.

\paragraph{Annotation scope.}
All metrics use each benchmark's provided boxes and temporal labels.
These standard protocols assess our amodal corrections but do not imply
a uniform GT convention for full object extent under occlusion.

\begin{figure*}[!t]
    \centering
    \includegraphics[width=\textwidth]{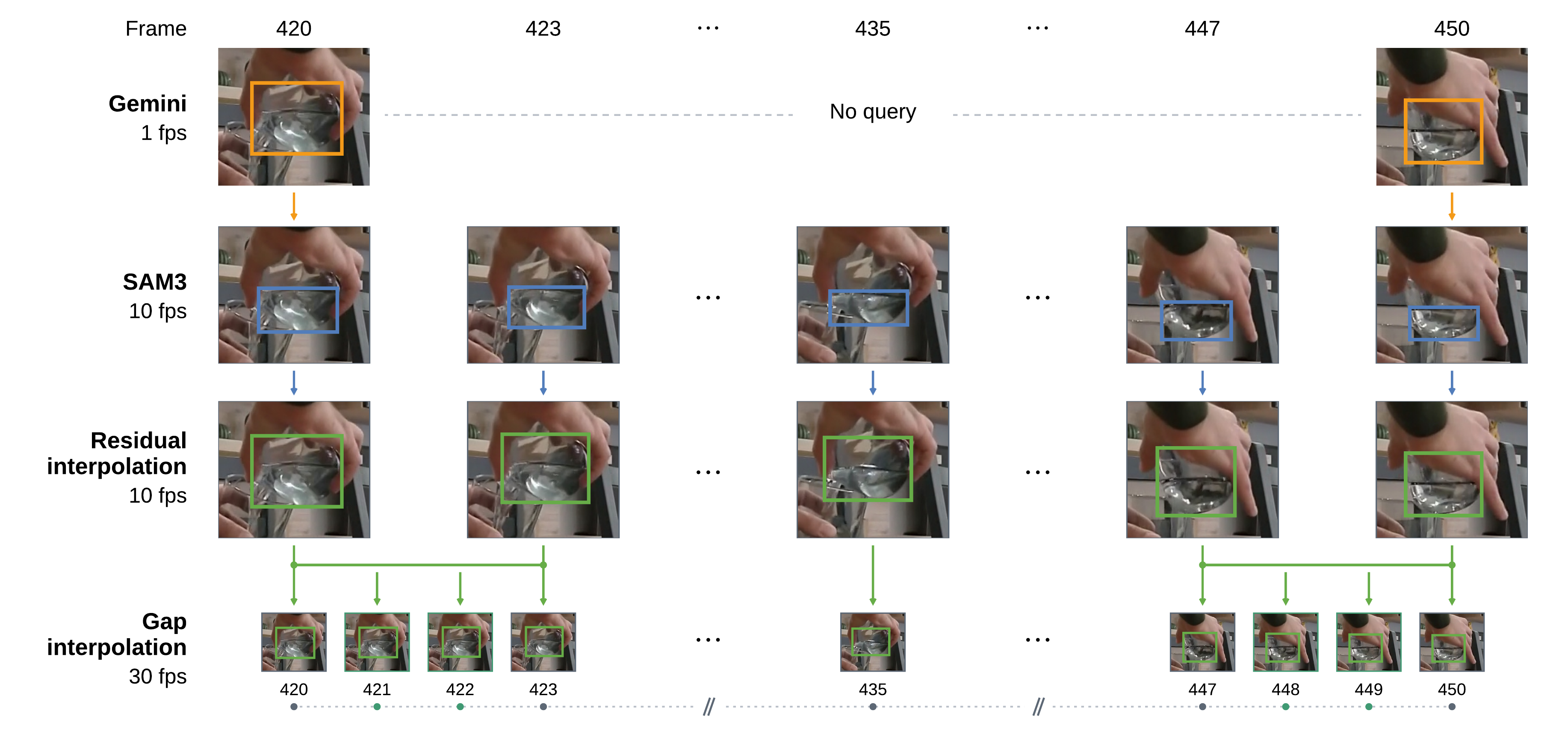}
    \caption{Sparse-to-dense fusion on a Perception Test example with a
    nominal 30-fps source grid. Gemini provides sparse amodal boxes at
    1\,fps (orange), and SAM3 provides visual tracks at 10\,fps (blue).
    Residual interpolation corrects existing SAM3 boxes (green), including
    frames without Gemini observations. Gap interpolation fills eligible
    missing frames on the original grid. Both endpoint observations are
    available for this offline interpolation.}
    \label{fig:supp-fusion-interpolation}
\end{figure*}

\section{Residual and Gap Interpolation}
\label{sec:supp-fusion-interpolation}

Figure~\ref{fig:supp-fusion-interpolation} illustrates the two interpolation steps in SD-Track. Residual interpolation corrects boxes at existing tracking frames, while gap interpolation creates boxes at missing source frames.

\paragraph{Residual interpolation.}
Gemini provides sparse amodal observations at frames 420 and 450. At shared frames where the sparse amodal and dense visual boxes satisfy the overlap criterion, we compute their coordinate differences. We linearly interpolate these residuals and add them to the existing SAM3 boxes. Frames 423, 435, and 447 therefore receive amodal corrections without additional Gemini queries. This combines SAM3's dense temporal tracking with Gemini's estimates of hidden object extent.

\paragraph{Gap interpolation.}
After correction and merging, we fill missing source frames between consecutive observations when their endpoint boxes have an IoU of at least $\tau=0.3$. In the illustrated interval, this adds boxes at frames 421 and 422 between frames 420 and 423, and at frames 448 and 449 between frames 447 and 450. These boxes are obtained by linearly interpolating the corrected endpoint coordinates. This step represents the trajectory on the original 30-fps frame grid without additional model inference. Gaps that fail the overlap criterion remain unfilled.

% Insert this section into the paper; requires amsmath, graphicx and booktabs.

% Float group B: threshold analysis figures following Section B.
\begin{table}[!t]
    \centering
    \caption{\textbf{Shared-threshold sweep on Perception Test val200.}
    All rows use 200 queries from 190 videos. Both thresholds vary
    together, and loss is measured against the observed maximum.
    All values are on the 0--100 HOTA scale.}
    \label{tab:val200_shared_tau}
    \footnotesize
    \setlength{\tabcolsep}{3pt}
    \renewcommand{\arraystretch}{1.12}
    \begin{tabularx}{\columnwidth}{@{}>{\raggedright\arraybackslash}Xrr@{}}
        \toprule
        \textbf{Shared $\tau_r=\tau_g$} & \textbf{HOTA} & \textbf{Loss to max.} \\
        \midrule
        0                     & 69.400 & 2.449 \\
        0.201 (observed max.) & 71.849 & 0.000 \\
        0.3 (reported)        & 71.802 & 0.048 \\
        1                     & 70.439 & 1.410 \\
        \bottomrule
    \end{tabularx}
\end{table}

\begin{figure*}[!t]
    \centering
    \includegraphics[width=0.87\textwidth]{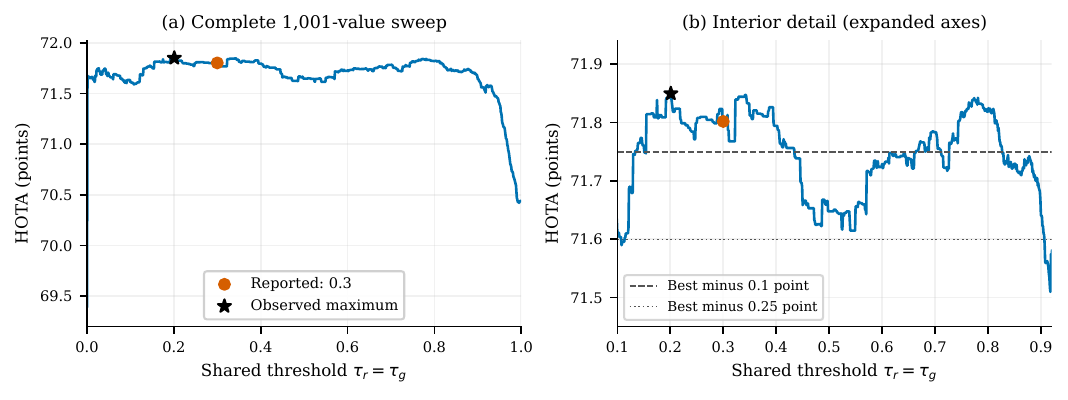}
    \caption{Shared-threshold sensitivity on native Perception Test val200.
    (a) The complete 1,001-value curve.
    (b) The interior region with expanded axes; horizontal lines mark
    descriptive losses of 0.1 and 0.25 HOTA points from the observed
    maximum. The reported value 0.3 is 0.048 points below that maximum.
    Both panels use the same stored evaluations.}
    \label{fig:val200_shared_tau}
\end{figure*}

\begin{figure*}[!tp]
    \centering
    \includegraphics[width=\textwidth]{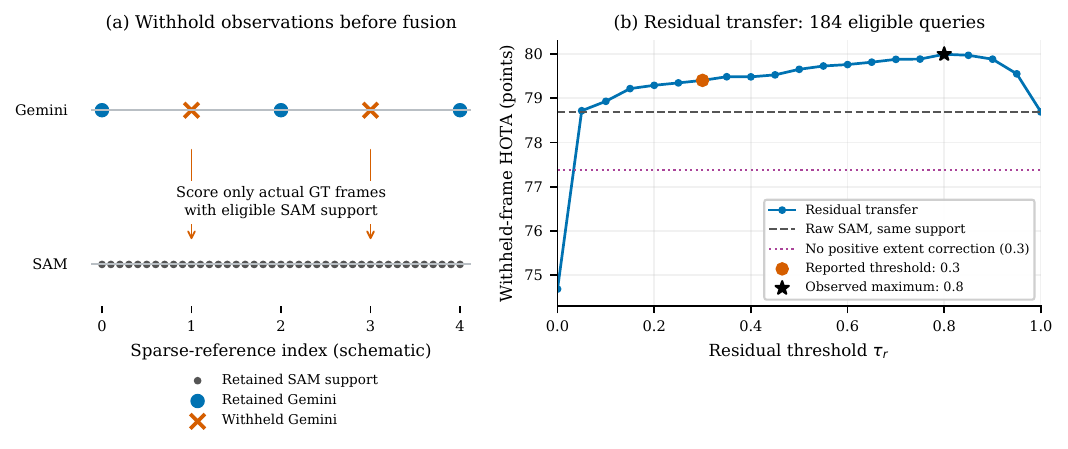}
    \caption{\textbf{Residual transfer with withheld Gemini observations.}
    (a) We remove alternating Gemini observations while keeping the underlying SAM3 trajectory fixed. The two runs use opposite groups
    of observations, and only eligible frames with original GT
    annotations are evaluated.
    (b) HOTA as the residual threshold $\tau_r$ varies on the
    same 184-query evaluation set. Gap interpolation and sparse
    fallback are disabled. Dashed and dotted lines indicate raw
    SAM3 and the variant without positive extent correction.
    The reported threshold of 0.3 improves over raw SAM3, while
    the highest observed HOTA occurs at 0.8.}
    \label{fig:val200_residual_holdout}
\end{figure*}

\begin{table}[!t]
    \centering
    \caption{\textbf{Residual transfer with withheld Gemini observations.}
    Results on 184 queries from 178 videos in the Perception Test
    validation set. Odd- and even-numbered Gemini query frames
    are alternately withheld to evaluate residual transfer at
    frames without direct Gemini observations.
    Gap interpolation and sparse fallback are disabled.
    }
    \label{tab:val200_residual_holdout}
    \footnotesize
    \setlength{\tabcolsep}{3pt}
    \renewcommand{\arraystretch}{1.12}
    \begin{tabularx}{\linewidth}{@{}>{\raggedright\arraybackslash}Xrr@{}}
        \toprule
        \textbf{Method} & \textbf{HOTA} & \textbf{$\Delta$ vs. SAM} \\
        \midrule
        Raw SAM & 78.690 & +0.000 \\
        Residual, $\tau_r=0$ & 74.686 & $-4.004$ \\
        Residual, $\tau_r=0.3$ (reported) & 79.405 & +0.715 \\
        Residual, $\tau_r=0.8$ (observed max.) & 79.996 & +1.305 \\
        Residual, $\tau_r=1$ & 78.690 & +0.000 \\
        No positive extent correction ($\tau_r=0.3$) & 77.370 & $-1.321$ \\
        \bottomrule
    \end{tabularx}
\end{table}

\begin{table*}[!t]
    \centering
    \caption{Localization by external occlusion on all 200 validation questions. $n$ counts question-conditioned GT object observations after uncertain and image-truncated observations are excluded. Scores are on a 0--100 scale. Unmatched GT objects receive zero IoU. Bold marks the best value for each metric in a row.}
\label{tab:supp-occ-degree}
    \footnotesize
    \setlength{\tabcolsep}{3pt}
    \renewcommand{\arraystretch}{1.10}
    \begin{adjustbox}{max width=\textwidth}
\begin{tabular}{lr*{4}{rr}}
\toprule
Occlusion & $n$ & \multicolumn{2}{c}{Dense SAM3} & \multicolumn{2}{c}{Sparse Gemini + linear} & \multicolumn{2}{c}{SD-Track} & \multicolumn{2}{c}{Direct Gemini} \\
\cmidrule(lr){3-4}\cmidrule(lr){5-6}\cmidrule(lr){7-8}\cmidrule(lr){9-10}
 & & mIoU $\uparrow$ & $R_{0.5}$ $\uparrow$ & mIoU $\uparrow$ & $R_{0.5}$ $\uparrow$ & mIoU $\uparrow$ & $R_{0.5}$ $\uparrow$ & mIoU $\uparrow$ & $R_{0.5}$ $\uparrow$ \\
\midrule
None & 3,475 & 83.19 & 93.27 & 82.03 & 90.33 & 85.32 & 94.13 & \textbf{85.43} & \textbf{95.28} \\
Partial & 2,270 & 71.13 & 81.10 & 74.55 & 84.89 & 74.74 & 84.58 & \textbf{76.87} & \textbf{87.89} \\
Full & 444 & 9.78 & 3.83 & \textbf{32.86} & \textbf{29.05} & 28.88 & 23.20 & 29.41 & 26.58 \\
\bottomrule
\end{tabular}
    \end{adjustbox}
\end{table*}

\begin{table*}[!t]
    \centering
    \caption{Complete-question tracking by occlusion group. Here $n$ counts questions, and scores are macro-averaged over complete question trajectories. Group membership uses confirmed observations without image truncation. Partial, no retained full means at least one retained partial observation and no retained full observation; uncertain or truncated observations may still occur elsewhere. Unresolved denotes questions with uncertain or truncated observations and no retained partial or full observation. All four groups sum to 200 questions.}
\label{tab:supp-occ-tracking}
    \footnotesize
    \setlength{\tabcolsep}{3pt}
    \renewcommand{\arraystretch}{1.10}
    \begin{adjustbox}{max width=\textwidth}
\begin{tabular}{lr*{4}{rr}}
\toprule
Question group & $n$ & \multicolumn{2}{c}{Dense SAM3} & \multicolumn{2}{c}{Sparse Gemini + linear} & \multicolumn{2}{c}{SD-Track} & \multicolumn{2}{c}{Direct Gemini} \\
\cmidrule(lr){3-4}\cmidrule(lr){5-6}\cmidrule(lr){7-8}\cmidrule(lr){9-10}
 & & HOTA $\uparrow$ & AssA $\uparrow$ & HOTA $\uparrow$ & AssA $\uparrow$ & HOTA $\uparrow$ & AssA $\uparrow$ & HOTA $\uparrow$ & AssA $\uparrow$ \\
\midrule
No external occlusion & 5 & 85.79 & 85.79 & 82.57 & 82.57 & \textbf{87.45} & \textbf{87.45} & 80.94 & 80.94 \\
Partial, no retained full & 133 & 75.02 & 76.63 & 70.29 & 70.42 & 78.82 & \textbf{80.27} & \textbf{78.95} & 79.78 \\
Contains full & 48 & 40.35 & 41.84 & 49.05 & 48.83 & 50.49 & 51.75 & \textbf{53.17} & \textbf{53.82} \\
Unresolved & 14 & 73.48 & 73.53 & 71.86 & 71.93 & 72.61 & 72.67 & \textbf{75.23} & \textbf{75.41} \\
\bottomrule
\end{tabular}
    \end{adjustbox}
\end{table*}

\begin{figure*}[!t]
\centering
\begin{minipage}[t]{0.34\textwidth}
    \vspace{0pt}
    \centering
    \suppOcclusionPlot{\linewidth}{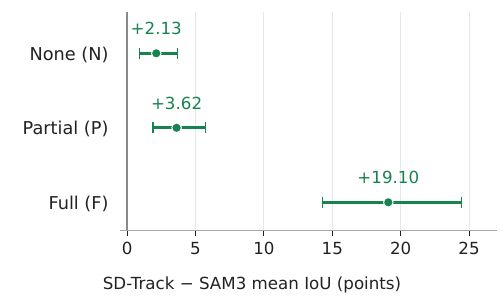}
    \caption{SD-Track mean-IoU gains over dense SAM3 by occlusion degree. Points show paired differences in IoU points, and bars show 95\% video-cluster bootstrap confidence intervals. Uncertain and image-truncated observations are excluded.}
    \label{fig:supp-occ-gains}
\end{minipage}\hfill
\begin{minipage}[t]{0.63\textwidth}
    \vspace{0pt}
    \centering
    \suppOcclusionPlot{\linewidth}{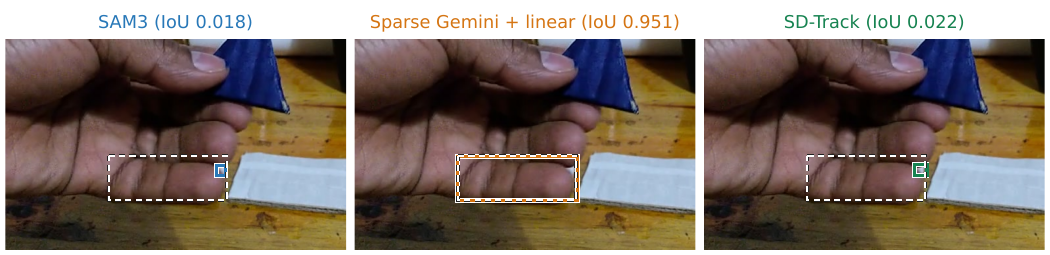}
    \caption{A selected full-occlusion failure on Perception Test
    validation. A hand hides the white rectangle at frame 480 of
    \texttt{video\_11451\_q3}. The accurate sparse box is rejected as a
    residual reference because the SAM3 box has collapsed. Dense-priority
    merging retains the small corrected box. Dashed boxes denote GT, and
    solid boxes denote each method's prediction. Values are frame-level IoU.}
    \label{fig:supp-occ-failure}
\end{minipage}
\end{figure*}

\section{Threshold Analysis and Residual Transfer}
\label{sec:val200_tau}

\subsection{The two IoU gates}
We use $\tau_r=\tau_g=0.3$ in the reported configuration. For a given target, let $S_t$ and $G_t$ denote its SAM and Gemini boxes at reference frame $t$, and let $B_a,B_b$ be available output boxes at two different frames. The residual gate accepts coordinate-correction references according to
\begin{equation}
\operatorname{IoU}(S_t,G_t)\geq\tau_r,
\end{equation}
whereas the gap gate admits interpolation between available endpoints when
\begin{equation}
\operatorname{IoU}(B_a,B_b)\geq\tau_g.
\end{equation}
The former measures agreement between models at the same time; the latter measures agreement across time. A common numerical setting therefore does not imply a common optimum. We distinguish sensitivity to the tied setting from a controlled test of residual transfer.

\subsection{Sensitivity to a common threshold}
We reuse the complete sweep $\tau_r=\tau_g\in\{0,0.001,\ldots,1\}$ on 200 Perception Test validation queries from 190 videos. Figure~\ref{fig:val200_shared_tau} and Table~\ref{tab:val200_shared_tau} show that 0.3 gives HOTA 71.802, while the observed maximum is 71.849 at 0.201, a difference of 0.048 points. The connected intervals within 0.1 and 0.25 points of this maximum are $[0.155,0.434]$ and $[0.114,0.906]$, respectively. Both endpoints perform worse: HOTA is 69.400 at zero and 70.439 at one. These small interior differences support retaining 0.3 as a common operating value. The intervals are descriptive tolerances, not statistical equivalence margins. This tied sweep changes both gates and cannot attribute its differences to either gate individually.

\subsection{Residual Transfer with Withheld Gemini Observations}
\label{sec:val200_residual_holdout}

In Perception Test, 3,854 of the 4,139 annotated GT query-frame
slots (93.114\%) coincide with Gemini query frames.
Evaluation therefore mostly covers times at which Gemini is
queried, giving limited evidence about how well its corrections
transfer to other frames.
We test this transfer by withholding saved Gemini boxes and
comparing the resulting SAM3 predictions with the original GT
at the withheld frames.

For each video, we sort the queried Gemini frames chronologically
and divide them into odd-numbered query frames (1st, 3rd, 5th, ...)
and even-numbered query frames (2nd, 4th, 6th, ...).
All Gemini boxes from the same frame are assigned to the same group.
In the first run, we retain the odd-numbered frames and withhold
the even-numbered frames. In the second run, we reverse this
selection, retaining the even-numbered frames and withholding
the odd-numbered frames.
This interleaving creates withheld interior frames between
retained observations and reduces the Gemini observation rate
supplied to fusion from approximately 1\,fps to 0.5\,fps.
Swapping the groups lets us evaluate eligible frames from both
groups without supplying their own Gemini boxes to fusion.

Both runs reuse predictions generated from the original video
context; we change only the saved Gemini boxes supplied to fusion.
At each retained reference, we compare the Gemini box with the
SAM3 box at the same time.
If their IoU is at least $\tau_r$, their coordinate difference
supplies a correction residual.
We interpolate these residuals over time and add them to the
SAM3 boxes at withheld frames.
For example, if references $t_0$ and $t_2$ are retained and
accepted, their Gemini--SAM3 differences correct the SAM3 box
at the withheld frame $t_1$.
The Gemini box at $t_1$ is absent throughout this calculation;
the corrected box is evaluated against the original GT at $t_1$.
Gap interpolation and sparse fallback are disabled, so the
experiment isolates corrections to existing SAM3 predictions.

We evaluate only withheld frames where SAM3 provides a prediction and both earlier and later frames have predictions from both SAM3 and Gemini. These evaluation frames are fixed before varying the residual threshold $\tau_r$.
After the two runs, each scored frame contributes the prediction
from the run in which its Gemini boxes were withheld.
We combine these predictions in temporal order and compute HOTA
once per query.
Raw SAM3 is evaluated on exactly the same frames as the baseline.
The resulting evaluation set contains 184 queries from 178 videos
and 2,958 GT query-frame slots.

Figure~\ref{fig:val200_residual_holdout}(a) illustrates the split,
and panel~(b) plots HOTA over the same evaluation set as $\tau_r$
varies.
Table~\ref{tab:val200_residual_holdout} lists selected points from
this curve and the comparison controls.
At the reported threshold $\tau_r=0.3$, residual transfer raises
HOTA from 78.690 to 79.405, a gain of 0.715 points
(95\% CI: $[0.231,1.166]$).
Accepting all references ($\tau_r=0$) instead lowers HOTA to
74.686, showing that transferring unreliable corrections can
harm the trajectory.
The highest observed HOTA is 79.996 at $\tau_r=0.8$.
This peak applies to the residual-only diagnostic with 0.5-fps
Gemini observations; the full pipeline uses approximately
1-fps observations together with gap interpolation.

To test whether box expansion contributes to the improvement,
we remove positive width and height corrections while preserving
center shifts, size reductions, and the accepted references at
$\tau_r=0.3$.
HOTA then falls from 79.405 to 77.370, a difference of 2.035 points
(95\% CI: $[1.729,2.358]$).
On this evaluation set, residual interpolation therefore
transfers useful size-expansion information from retained Gemini
observations to existing SAM3 boxes at frames that receive no
Gemini box during fusion.

\section{Prompt Design and Rationale}
\label{sec:rationale}

The planner first identifies the objects that answer the question. A second call checks that choice against the video. The planner then assigns each verified instance an ID and selects anchor frames where the instances can be distinguished (Figure~\ref{fig:prompts_plan}).

At each anchor, SAM3 starts from a box around the visible part of the target. At sparse frames, Gemini uses the same ID and neighboring frames to estimate a box around the entire object, including any hidden part. It keeps a fully occluded target only when the surrounding frames support its presence and omits off-screen targets (Figure~\ref{fig:prompts_localizing}).

\section{Occlusion analysis on Perception Test}
\label{sec:supp-occlusion}

\begin{figure*}[p]
\noindent\begin{minipage}[t][0.90\textheight][t]{0.485\textwidth}
\vspace{0pt}\raggedright\promptfont\fontsize{9}{10.4}\selectfont
\setlength{\parskip}{1pt plus .12fill}
{\setlength{\parskip}{0pt}\color{black}\rule{\linewidth}{0.7pt}\par\medskip\fontsize{13}{15}\selectfont\bfseries Planning\par}
\promptstep{Identify the target set}{Labeled planning video}
Watch the labeled video and answer \{query\}. Identify the physical instance or set of instances that satisfies the query. Use the action or relationship in the video to distinguish the target from other participants.

Return one concise noun phrase with enough visible detail to recognize the target. Leave frame selection and localization to the later steps.

Call submit\_\allowbreak{}answer\_\allowbreak{}noun\_\allowbreak{}phrase exactly once.
\promptreturn{Return}{answer\_\allowbreak{}noun\_\allowbreak{}phrase, reasoning, confidence}

\promptstep{Verify the target set}{Labeled planning video}
Work out which target satisfies \{query\}, then compare your answer with \{candidate\}. Check who acts, what is acted on, and any temporal or counterfactual condition in the query.

Count the distinct target instances. Return correct if the candidate identifies the set. Otherwise return wrong, give the corrected noun phrase, and cite the original frame labels that support it.

Call submit\_\allowbreak{}answer\_\allowbreak{}verification exactly once.
\promptreturn{Return}{reasoning, verdict, expected\_\allowbreak{}instance\_\allowbreak{}count. On correction: corrected\_\allowbreak{}noun\_\allowbreak{}phrase, evidence\_\allowbreak{}frames.}

\promptstep{Assign IDs and select anchors}{Labeled planning video}
Use \{noun\_\allowbreak{}phrase\} as the verified answer to \{query\}. Give each instance a stable obj\_\allowbreak{}id starting at 0, a label, and distinguishing visual features. Keep the same ID when it reappears. Account for \{count\_\allowbreak{}hint\} when supplied.

For each ID, list all visible intervals as inclusive pairs of original frame indices. Split intervals when the object disappears. Report the first and last visible frames of the whole set and the evidence for these bounds.

Choose one frame from \{allowed\_\allowbreak{}frames\} where all targets are visible, separated, and minimally occluded. If necessary, choose the smallest set of up to \{max\_\allowbreak{}anchor\_\allowbreak{}frames\} frames covering all targets. List the IDs visible in each chosen frame and explain the choice.

Call submit\_\allowbreak{}multi\_\allowbreak{}anchor\_\allowbreak{}selection exactly once.
\promptreturn{Return}{decision\_\allowbreak{}summary, first\_\allowbreak{}visible\_\allowbreak{}frame\_\allowbreak{}idx, last\_\allowbreak{}visible\_\allowbreak{}frame\_\allowbreak{}idx, visibility\_\allowbreak{}boundary\_\allowbreak{}reason, targets[], anchors[]. targets: obj\_\allowbreak{}id, object\_\allowbreak{}label, visual\_\allowbreak{}identity\_\allowbreak{}description, visibility\_\allowbreak{}segments[\{first\_\allowbreak{}frame\_\allowbreak{}idx, last\_\allowbreak{}frame\_\allowbreak{}idx\}]. anchors: frame\_\allowbreak{}idx, obj\_\allowbreak{}ids, reason.}
\par\vfill{\color{black}\rule{\linewidth}{0.4pt}}\par
\normalfont\normalsize\caption{Planning prompts in SD-Track.}
\label{fig:prompts_plan}
\end{minipage}\hfill
\begin{minipage}[t][0.90\textheight][t]{0.485\textwidth}
\vspace{0pt}\raggedright\promptfont\fontsize{9}{10.4}\selectfont
\setlength{\parskip}{1pt plus .12fill}
{\setlength{\parskip}{0pt}\color{black}\rule{\linewidth}{0.7pt}\par\medskip\fontsize{13}{15}\selectfont\bfseries Localization\par}
\promptstep{Draw visible boxes at anchor frames}{Planning video and exact anchor image}
The verified target for \{query\} is \{noun\_\allowbreak{}phrase\}. In the exact source image at \{frame\_\allowbreak{}idx\}, locate every ID in \{obj\_\allowbreak{}ids\} using \{identity\_\allowbreak{}descriptions\}. Keep each ID unchanged.

Draw one tight box around the visible extent of each requested instance. Exclude cast shadows and surrounding occluders. Return positive-area integer boxes in [ymin, xmin, ymax, xmax] order, normalized to 0–1000.

Call submit\_\allowbreak{}multi\_\allowbreak{}anchor\_\allowbreak{}boxes exactly once.
\promptreturn{Return}{boxes[\{obj\_\allowbreak{}id, box\_\allowbreak{}2d, reason, confidence\}]}

\par
\promptstep{Estimate amodal boxes across frames}{Ordered exact source images}
Process the \{width\} \texttimes{} \{height\} source images in the order given by \{frame\_\allowbreak{}indices\}. Keep the identities in \{compact\_\allowbreak{}plan\}. Use each target's own pixels, scale, and motion across the sequence to infer its full physical extent.

Include every ID listed in \{visibly\_\allowbreak{}confirmed\_\allowbreak{}ids\_\allowbreak{}by\_\allowbreak{}frame\}. During a gap in a visible interval, retain a hidden target when the surrounding frames support its presence. Omit targets that have left the image. Return one record per frame and an empty objects list when none remain.

Use visible when no substantial part is hidden. Use partially\_\allowbreak{}occluded when a substantial part is hidden and some target pixels remain. Use fully\_\allowbreak{}occluded when no pixels are visible and the sequence supports an in-frame location. Contact alone does not establish occlusion.

Return a separate amodal box for each present ID in the same integer coordinate format as the anchor call. Check each box against the target's scale. Exclude the occluder's extent and avoid copying a group box across IDs.
\promptreturn{JSON response}{frames[\{frame\_\allowbreak{}idx, objects[\{object\_\allowbreak{}id, visibility, box\_\allowbreak{}2d, confidence\}]\}]}
\par\vfill{\color{black}\rule{\linewidth}{0.4pt}}\par
\normalfont\normalsize\caption{Localization prompts in SD-Track.}
\label{fig:prompts_localizing}
\end{minipage}
\end{figure*}

\subsection{Annotation and evaluation protocol}
We analyze all 200 questions from 190 videos in the Perception Test validation subset~\cite{Patraucean2023PerceptionTest}. The answers contain 342 distinct GT object tracks and 6,826 distinct annotated object observations. A shared object can answer more than one question, so evaluating the question-conditioned predictions yields 6,932 observations. We reuse the saved outputs of dense SAM3, sparse Gemini with linear interpolation, direct Gemini, and SD-Track. Sparse Gemini uses the same 1-fps amodal observations as SD-Track. Direct Gemini predicts amodal boxes at every source frame. All trajectories use the original video frame grid.

\paragraph{Occlusion labels.}
Codex assistants assigned labels from source frames with only GT boxes overlaid; predictions and method scores were not displayed. Labels were shared across questions referring to the same object in the same video. We distinguish \emph{none}, where the object has no external occlusion; \emph{partial}, where another object hides some of the target but part remains visible; and \emph{full}, where the target is entirely hidden but video context indicates that it remains in the scene. Pose changes and self-occlusion do not count as external occlusion. We separately record image truncation and mark uncertain visibility or target identity as unresolved. These are coarse visual categories, without measured visible-area ratios.

The primary frame analysis excludes unresolved and image-truncated observations, leaving 6,189 question-conditioned observations. A separate AI review covered 361 distinct object observations and produced 395 paired label comparisons. Agreement on occlusion degree was 93.16\%, with Cohen's $\kappa=0.887$. Both reviews used the same model family. The labels have not been verified by human annotators, so we treat this as an exploratory validation analysis.

\paragraph{Metrics.}
At each annotated frame, we match all predicted objects to all answer objects before selecting an occlusion category. Mean IoU uses a one-to-one assignment that maximizes total IoU, with unmatched GT objects assigned zero. Recall at IoU 0.5, denoted $R_{0.5}$, uses a separate assignment that first maximizes the number of matches passing the threshold. Frame metrics are averaged over question-conditioned object observations. They measure localization and detection coverage. We evaluate temporal association separately using HOTA and AssA~\cite{qtrace_hota} on complete question trajectories. Paired 95\% confidence intervals use video-cluster percentile bootstrap with 3,000 resamples, keeping observations from the same video together. These intervals condition on the assigned labels and do not quantify annotation uncertainty.

\subsection{Localization under different occlusions}
% Inlined from occlusion_degree.tex (single-file supplementary)

Table~\ref{tab:supp-occ-degree} reports localization by occlusion degree. Under partial occlusion, SD-Track raises mean IoU from 71.13 to 74.74 relative to SAM3, a gain of 3.62 points (95\% CI: $[1.89,5.73]$). Recall also rises from 81.10 to 84.58. Sparse Gemini with interpolation reaches 74.55 mean IoU, close to SD-Track, while direct Gemini obtains the highest partial-occlusion score of 76.87. Thus, the frame results establish a gain over visible-mask tracking but do not establish a gain over sparse amodal localization.

Figure~\ref{fig:supp-occ-gains} shows the mean-IoU gains over SAM3 and their paired confidence intervals for each occlusion degree.

Full occlusion is more difficult. SD-Track improves mean IoU over SAM3 from 9.78 to 28.88, but sparse Gemini with interpolation reaches 32.86. The SD-Track minus sparse difference is $-3.98$ points (95\% CI: $[-7.65,-1.66]$). Sparse amodal predictions can therefore be useful even when the fusion rule does not retain their full benefit.

\subsection{Tracking over complete trajectories}
% Inlined from complete_trajectory.tex (single-file supplementary)

For Table~\ref{tab:supp-occ-tracking}, we group questions using confirmed labels on observations without image truncation. We then evaluate each complete trajectory with all its original evaluation frames. This keeps association across time in the evaluation. The partial group contains 133 questions with a retained partial observation and no retained full observation. Uncertain or truncated full observations can still occur in these questions.

On this group, SD-Track obtains 78.82 HOTA, compared with 75.02 for SAM3 and 70.29 for sparse Gemini with interpolation. Its gain over sparse is 8.53 points (95\% CI: $[6.65,10.54]$), and AssA increases from 70.42 to 80.27. The separate partial-observation analysis shows similar frame-level IoU for SD-Track and sparse Gemini. On these 133 complete trajectories, however, combining the sparse observations with dense tracking yields higher HOTA and AssA. Direct Gemini obtains 78.95 HOTA on the same questions. For the 48 questions containing a retained full-occlusion observation, SD-Track obtains 50.49 HOTA versus 49.05 for sparse Gemini; the paired difference of 1.44 points has a confidence interval of $[-1.21,4.54]$.

% Queue the full-width failure example near its discussion.

\subsection{A failure case under full occlusion}
\label{sec:supp-occ-failure}
Figure~\ref{fig:supp-occ-failure} shows a selected failure at frame 480 of \texttt{video\_11451\_q3}. A hand fully hides the white rectangular target. The sparse box has IoU 0.951 with GT, while SAM3 leaves a small box with IoU 0.018. Their cross-branch IoU falls below $\tau=0.3$, so the sparse observation is excluded from the correction references. Interpolating the nearby residuals only slightly changes the collapsed SAM3 box. Dense-priority merging then overwrites the accurate sparse box, leaving a fused IoU of 0.022. Reconstructing this calculation reproduces the saved fusion coordinates exactly. This case identifies a failure of the overlap check together with dense-priority merging; it does not measure how often the failure occurs.

Hidden-object GT locations can also be approximate. These diagnostics assess overlap with those boxes and do not isolate the contribution of boundary expansion from changes in position or detection coverage. The validation subset was used during method development, and the AI-generated occlusion labels require human verification before they can be treated as validated annotations.

\FloatBarrier

\end{document}